\documentclass{article} % For LaTeX2e
\usepackage{iclr2024_conference,times}

\usepackage{amsmath,amsfonts,bm}

\def\eqref#1{equation~\ref{#1}}
\def\1{\bm{1}}

\def\vzero{{\bm{0}}}

\def\vtheta{{\bm{\theta}}}
\def\va{{\bm{a}}}

\def\ve{{\bm{e}}}

\def\vg{{\bm{g}}}

\def\vs{{\bm{s}}}

\def\vu{{\bm{u}}}

\def\vy{{\bm{y}}}

\def\mA{{\bm{A}}}

\def\mE{{\bm{E}}}
\def\mF{{\bm{F}}}
\def\mG{{\bm{G}}}
\def\mH{{\bm{H}}}
\def\mI{{\bm{I}}}

\def\mK{{\bm{K}}}

\def\mS{{\bm{S}}}

\def\mU{{\bm{U}}}
\def\mV{{\bm{V}}}
\def\mW{{\bm{W}}}

\DeclareMathAlphabet{\mathsfit}{\encodingdefault}{\sfdefault}{m}{sl}
\SetMathAlphabet{\mathsfit}{bold}{\encodingdefault}{\sfdefault}{bx}{n}

\newcommand{\E}{\mathbb{E}}

\newcommand{\R}{\mathbb{R}}

\DeclareMathOperator*{\argmin}{arg\,min}

\usepackage{hyperref}
\usepackage{url}
\usepackage{graphicx}
\usepackage{cleveref}
\usepackage{mathtools,amsmath}

\usepackage{wrapfig}

\usepackage{algorithm}
\usepackage{algpseudocode}

\newcommand*\widebar[1]{%
  \hbox{%
    \vbox{%
      \hrule height 0.5pt % The actual bar
      \kern0.5ex%         % Distance between bar and symbol
      \hbox{%
        \kern-0.1em%      % Shortening on the left side
        \ensuremath{#1}%
        \kern-0.1em%      % Shortening on the right side
      }%
    }%
  }%
} 

\title{\vspace{-1em}The LLM Surgeon}

\author{Tycho F.A. van der Ouderaa$^{1}$\thanks{Work done while doing an internship at Qualcomm AI Research}\ \ , Markus Nagel$^{2}$, Mart van Baalen$^{2}$, \\ 
\textbf{Yuki M. Asano$^{3}$, Tijmen Blankevoort$^{2}$} \\
$^{1}$Imperial College London\ , $^{2}$Qualcomm AI Research\thanks{Qualcomm AI Research is an initiative of Qualcomm Technologies, Inc. }\ , $^{3}$QUVA Lab,
University of Amsterdam
}

\iclrfinalcopy % Uncomment for camera-ready version, but NOT for submission.

\begin{document}

\maketitle

\vspace{-1.9em}
\begin{abstract}
\vspace{-0.7em}
State-of-the-art language models are becoming increasingly large in an effort to achieve the highest performance on large corpora of available textual data. However, the sheer size of the Transformer architectures makes it difficult to deploy models within computational, environmental or device-specific constraints. We explore data-driven compression of existing pretrained models as an alternative to training smaller models from scratch. 
To do so, we scale Kronecker-factored curvature approximations of the target loss landscape to large language models. In doing so, we can compute both the dynamic allocation of structures that can be removed as well as updates of remaining weights that account for the removal. We provide a general framework for unstructured, semi-structured and structured pruning and improve upon weight updates to capture more correlations between weights, while remaining computationally efficient. 
Experimentally, our method can prune rows and columns from a range of OPT models and Llamav2-7B by 20\%-30\%, with a negligible loss in performance, and achieve state-of-the-art results in unstructured and semi-structured pruning of large language models. \\
Code is available at: \href{https://github.com/Qualcomm-AI-research/llm-surgeon}{https://github.com/Qualcomm-AI-research/llm-surgeon}.
\end{abstract}
\vspace{-2em}

\begin{figure}[b]
\vspace{-0.5em}
\centering
{\hfil Structured compression (rows and columns)\hfil Unstructured compression (matrix elements) \\
\vspace{0.5em}}
\resizebox{1.0\linewidth}{!}{
\includegraphics[]{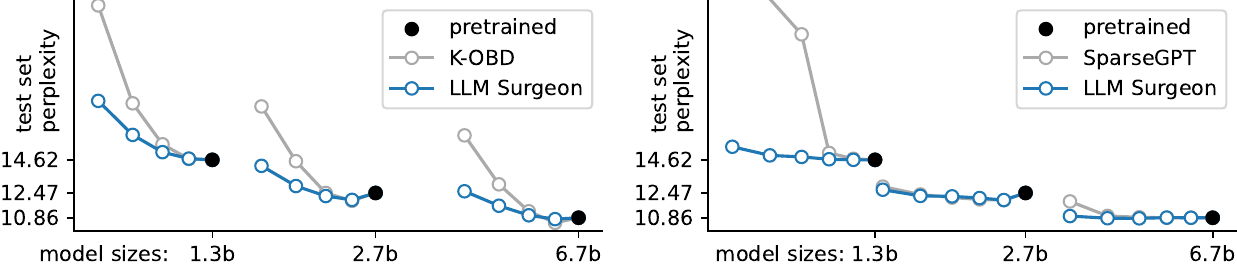}
}
\vspace{-1.7em}
\caption{LLM Surgeon allows interpolation of model size between existing pretrained models.}
\label{fig:horizontal-overview}
\end{figure}

\section{Introduction}
\label{introduction}
Recent advancements in language modeling \citep{vaswani2017attention} allow fitting large language models (LLMs) with millions or even billions of parameters (such as OPT \citep{zhang2022opt} and Llama~2 \citep{touvron2023llama}) on big text corpora achieving high performance. Unfortunately, the size of these LLMs often makes it hard to deploy them within practical constraints. Cloud-based deployment can get very expensive for larger models, and efficient devices such as phones are frequently limited in the memory size to host a model. 

% This work argues for data-based compression of existing large language models into smaller models to meet deployment constraints, as an alternative to training new models from scratch. The key benefits of this learning paradigm are (i) leveraging the performance of existing pretrained LLMs without requiring large datasets or the need for expensive training, while at the same time (ii) obtaining a model that exactly fits deployment requirements, including environmental or hardware constraints.

%  Compared to existing LLM pruning work, the main contributions of this work are the use of more accurate approximations of the loss landscape curvature and taking more weight correlations into account when updating remaining weights. Unlike most prior on data-based compression of LLMs, we not only use weight strengths and activations from forward passes, but also gradient information from backward passes to relate expected cost of weight removals to the final global objective, which allows for global thresholding. In addition, we explore multi-shot procedures and first-order weight corrections to further improve compression performance. Experimentally, the method outperforms state-of-the-art in large LLM pruning for unstructured, semi-structured and structured compression.

A body of literature extending back to the late 1980s, e.g., Optimal Brain Damage (OBD, \citet{lecun1989optimal}) and Optimal Brain Surgeon (OBS, \cite{hassibi1992second}), phrases pruning as a constraint optimization problem to reduce a model's footprint and runtime requirements.
The Hessian required for this approach grows with the square of the number of parameters, and can only be computed in practice for unrealistically small networks. To overcome this issue, Eigendamage \citep{wang2019eigendamage} introduces a Kronecker factorization of a blockwise-diagonal approximation of the Hessian. Recent works, like Optimal Brain Compression 
\citep{frantar2022optimal}, SparseGPT \citep{frantar2023sparsegpt}, demonstrate practical post-training pruning of LLMs, but only consider a loss curvature of a pruned layer's squared output reconstruction error, ignoring gradients that relate local removal costs to the target loss. As a result, their approximation to the target loss landscape is inaccurate, leading to a significant performance degradation for pruned LLMs. Further, these methods do not readily extend to structured pruning. 

This work introduces LLM Surgeon, a general framework for unstructured, semi-structured and structured pruning of LLMs. At paper submission, we deemed this the first method to successfully perform structured pruning of LLMs. Concurrent work by \cite{ashkboos2024slicegpt} also considers structured pruning of LLMs but ignores gradient information, resulting in lower final performance. The superior performance of LLM Surgeon is achieved by scaling up the block-diagonal Kronecker-factorized approximations to the empirical Fisher from Eigendamage to LLMs.
We further expand upon the work by deriving OBS-like weight pruning costs and updates for structured pruning of multiple rows and columns, and provide a general framework that also incorporates semi-structured and unstructured pruning. Instead of treating individual weight updates independently, we strive to consider as many correlations between weights as practically possible and derive joint weight updates for pruning multiple weights (or multiple sets of structured weights) at once. Unlike prior work in LLM pruning, LLM Surgeon prunes in multiple shots, updating weights and curvature estimates between shots. We use global thresholding for unstructured, semi-structured and structured, i.e., instead of pruning layers by a fixed amount, more sensitive layers are pruned less than those that are more robust. Lastly, we propose to mitigate possible first-order gradients not being zero by using optional low-rank first-order updates between shots. 
A key advantage of LLM Surgeon is that it allows trading off additional compute during compression for better accuracy by increasing the number of correlations and/or shots. 
Our method gives the first practically usable results for structured pruning of LLMs -- they can be pruned by up to 30\% with minor performance degradation. Furthermore, we achieve state-of-the-art results in unstructured and semi-structured LLM pruning.

% This work introduces LLM Surgeon. Compared to existing LLM pruning work, LLM Surgeon uses a more accurate approximation of the loss landscape curvature and takes more weight correlations into account when updating unpruned weights. Unlike most prior work on data-based compression of LLMs, we not only use weight strengths and activations from forward passes, but also gradient information from backward passes to relate expected cost of weight removals to the final global objective. Using gradient information allows LLM Surgeon to use global thresholding, meaning layers that are less sensitive to pruning will be pruned more than those that are more sensitive. In addition, we explore multi-shot procedures and first-order weight corrections to further improve compression performance. Experimentally, the method shows state-of-the-art performance in large LLM pruning for unstructured, semi-structured and structured compression.

% %\newpage
\section{Background and related work}
\label{background}

% \begin{figure}[t]
% \centering
% \resizebox{0.95\linewidth}{!}{
% \includegraphics[]{figures/ad1.png}
% }
% \vspace{-1.0em}
% \caption{Instead of (a) training from scratch on a small amount of target data, we propose (b) data-based surgery of pretrained LLM. For example, for (c) on-device deployment of compressed model.}
% \vspace{-1em}
% \end{figure}

% \begin{figure}[t]
% \resizebox{0.9\linewidth}{!}{
% \includegraphics[]{figures/ad2.png}
% }
% \vspace{-1em}
% \caption{(a) Unstructure pruning has little practical improvements. (b) Structured pruning through column removal, row removal or lower rank representations achieves direct gain in memory and compute requirements. (c) Dynamic allocation allows reduction in layers of least important.}
% \end{figure}

Neural network pruning aims to remove parameters from a model while minimizing negative impact on final performance. More formally, we denote the $P$ model parameters as vector $\vtheta^* = \text{vec}(\mW^*_1, \mW^*_2, \ldots \mW^*_L) \in \R^P$, by flattening the $L$ weight matrices of attention and fully-connected blocks, with already fitted $\vtheta^* {\approx} \argmin_\vtheta \mathcal{L}(\vtheta)$ to data $\mathcal{D}$ to minimise a negative likelihood loss $\mathcal{L}(\vtheta) {=} -\log p(\vtheta | \mathcal{D})$. To compress the model, we are looking for a pruned vector $\hat{\vtheta}$:
\begin{align}
\hat{\vtheta} = \argmin\nolimits_\vtheta \mathcal{L}(\vtheta)\text{   s.t. pruning constraints based on } \vtheta^*
\label{eq:pruning-problem}
\end{align}
% \begin{align}
% \hat{\vtheta}, \hat{\vm} = \argmin_{\vtheta, \vm} \mathcal{L}(\vm \odot \vtheta)
% \label{eq:pruning-problem}
% \end{align}
where chosen constraints determine the structure of compressed weights $\hat{\vtheta}$.
In \textbf{unstructured~pruning}, a fraction of total weight elements is set to zero. In \textbf{semi-structured~pruning} of M:N we have that M weights of every N consecutive weights are zero \citep{zhou2021learning,hubara2021accurate}. And in \textbf{structured~pruning} \citep{louizos2017learning}, entire rows and columns are set to zero. Structured pruning leads to the most immediate gains in memory and computing, as it directly reduces the dimensions of matrices that need to be represented explicitly but is regarded as the most difficult to compress. Maintaining high performance is often easier in the other schemes but requires specialised arithmetic exploiting the sparsity structure to benefit at deployment. We consider all pruning types above, with a focus on structured pruning for LLMs.

Typically, \cref{eq:pruning-problem} can not be solved directly, as the space of possible pruning configurations exceeds what can be evaluated in practice. To illustrate, a search over all possible unstructured pruning masks of a 125 million parameter LLM would require $2^P{=}2^{125\text{m}}{\approx}10^{37628749}$ evaluations. The idea, therefore, is to find $\hat{\vtheta}$ using a surrogate of the loss landscape $q$ that is easier to work with:
\begin{align}
\mathcal{L}(\vtheta) = - \log p(\mathcal{D} \mid \vtheta) \approx  - \log q(\vtheta)
\end{align}
If one chooses a particular Gaussian form for our surrogate $q$, then solutions for unstructured, semi-structured, and structured pruning constraints can be derived in closed-form (\cref{sec:structured-pruning-derivations}).

\subsection{Taylor expansion}
\label{sec:taylor}
\begin{figure}
\resizebox{\linewidth}{!}{
\includegraphics[]{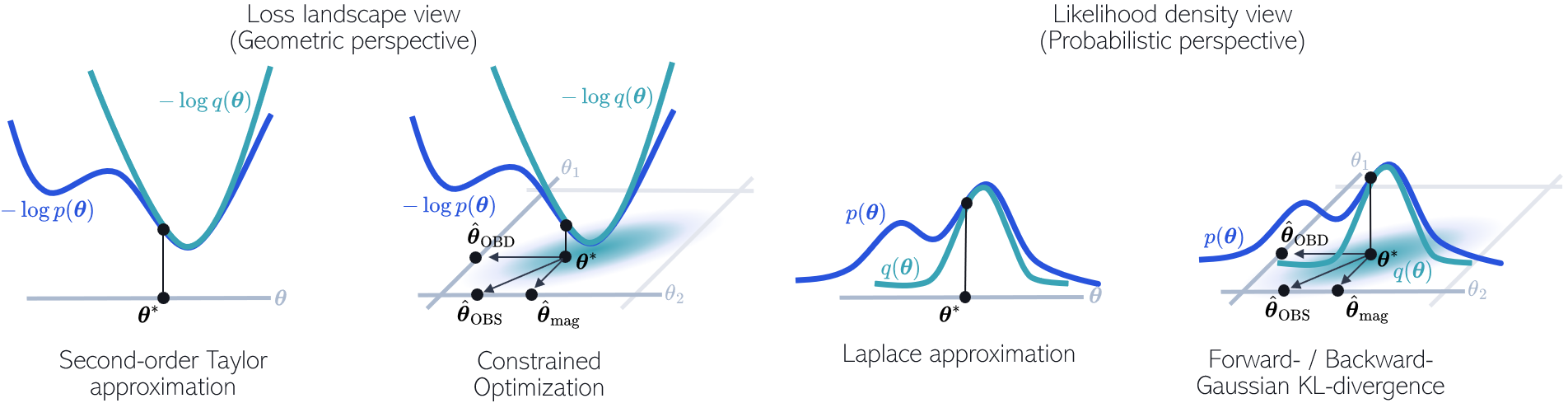}
}
\vspace{-2.5em}
\caption{Pruning as equality constrained optimization of quadratic approximation of the loss landscape (left), or equivalently, maximising the likelihood under a Laplace approximation (right).}
\vspace{-1em}
\label{fig:ad_geom_prob}
\end{figure}

How do we obtain a good surrogate of the loss $q$? One of the easiest approaches is to locally expand the log loss through a second-order Taylor expansion around the pretrained weights $\vtheta^*$, yielding:
\begin{align}
\label{eq:taylor}
-\log q(\vtheta) \approx -\log p(\mathcal{D} | \vtheta^*) - (\vtheta - \vtheta^*)^T \nabla \mathcal{L}(\vtheta^*) - \frac{1}{2} (\vtheta - \vtheta^*)^T \mH_{\vtheta^*} (\vtheta - \vtheta^*)
\end{align}
where $[\nabla \mathcal{L}(\vtheta^*)]_i = \frac{\partial}{\partial \vtheta_i} \mathcal{L}(\vtheta_i^*)$ denotes the Jacobian and $[\mH_{\vtheta}]_{ij} = \frac{\partial^2}{\partial \vtheta_i \vtheta_j} \mathcal{L}(\vtheta_{ij})$ denotes the Hessian. The first-order term vanishes $[\nabla \mathcal{L}(\vtheta^*)]_i = \vzero $ at the optimum. 
% This is clearly an oversimplification, since (i) the neural network may not be optimised to the minimum, (ii) we may use a different loss for compression than was used to train the network, (iii) we prune in multiple shots (see \cref{sec:multiple-shot}) inevitably causing weights to diverge from the optimum. 
% Nevertheless, we initially follow this simplifying assumption and consider interleaved first-order corrections to mitigate the issue in \cref{sec:interleaved-lora}. 
Note that in practice the first order term may not vanish.
While we follow this assumption initially, we consider interleaved first-order corrections to mitigate the issue in \cref{sec:interleaved-lora}. 
The quadratic expansion of \cref{eq:taylor} forms the basis of the optimal brain damage \citep{lecun1989optimal} and optimal brain surgeon \citep{hassibi1992second} pruning methods. Note that from a probabilistic perspective, a quadratic approximation of the log likelihood implies a Gaussian approximation of the likelihood, as also observed by \citep{wang2019eigendamage} and illustrated in \cref{fig:ad_geom_prob}. This is well-known \citep{bishop2006pattern}, \citep{mackay2003information} as the Laplace approximation $q(\vtheta) = \mathcal{N}(\vtheta \mid \vtheta^* + \nabla \mathcal{L}(\vtheta^*), \mH_{\vtheta^*}^{-1}$), with pretrained weights are the mean and the local inverse Hessian is the covariance matrix capturing \textit{correlations between weights}.

\subsection{Block Fisher Information Matrix}
\label{sec:block-fisher-information-matrix}

% A common way to approximate the Hessian is through the Fisher Information Matrix:
For a network trained with negative log-likehood loss, the Hessian is identical to the Fisher matrix: % Information Matrix (FIM):
\begin{align}
\label{eq:fim}
% \mH_{\vtheta} \approx \mF_{\vtheta} = \sum_{n=1}^N \E_{y\sim p_{\vtheta}(y | x_n)} \left[\nabla_\vtheta \log p_{\vtheta} (y | x_n)\nabla_\vtheta \log p_{\vtheta} (y | x_n)^T \right]
\mH_{\vtheta} = \mF_{\vtheta} = \sum\nolimits_{n=1}^N \E_{y\sim p_{\vtheta}(y | x_n)} \left[\nabla_\vtheta \log p_{\vtheta} (y | x_n)\nabla_\vtheta \log p_{\vtheta} (y | x_n)^T \right]
\end{align}
which has the benefit of always being positive semi-definite,
with the inverse thus forming a proper covariance matrix for $q$, 
and can be approximated with Monte Carlo samples of $p_{\vtheta}(y | x_n)$. For most LLMs, this would be treating the softmax output of the network as categorical distribution $p_{\vtheta}(y | x_n)$, and sampling from that. In practice, we use the `empirical Fisher' replacing the expectation over $y$ with target data $y_n$ \citep{kunstner2019limitations}. The full (empirical) Fisher $\mF_{\vtheta} \in \R^{P \times P}$ scales quadratically in the number of parameters $P$. To overcome this, the Fisher is often written in terms of layer-wise blocks $\mF_{lk} = \sum_{n=1}^N \E\left[ \text{vec}(\nabla_{\mW_l} \log p_{\vtheta}(y | x_n))  \text{vec}(\nabla_{\mW_k} \log p_{\vtheta}(y | x_n))^T \right]$, and approximated by only treating layers independently \citep{martens2015optimizing,botev2017practical}:
\begin{align}
\label{eq:block-fisher}
\mF_{\vtheta} =
\text{diag}(\mF_{11}, \mF_{22}, \ldots, \mF_{LL}), \hspace{3em}
\mF_{l}
&= \sum\nolimits_{n=1}^N  \E \Big[ \underbrace{(\vg_{l, n} \vg_{l, n}^T) \otimes (\va_{l, n} \va_{l, n}^T)}_{ RC \times RC }  \Big] 
\end{align}
where $\otimes$ denotes the Kronecker product and $\text{vec}(\cdot)$ the matrix vectorisation operation. Because we disregard cross-layer interactions we write $\mF_l$ instead of $\mF_{ll}$ for Fisher blocks associated with the weight matrix $\mW_l {\in} \R^{R \times C}$ producing outputs $\vy_{l, n} = \mW_l \va_{l, n} {\in} \R^{R}$ from inputs $\va_{l,n} {\in} \R^C$, for each layer $l$ and datapoint $n$. Consequently, we can compute Fisher blocks from input activations $\va_{l, n} {\in} \R^{C}$ of forward-passed data $x_n$ and output gradients $\vg_{l,n} {=} \nabla_{\vy_{l, n}} \mathcal{L} {\in} \R^{R}$ from backpropagation.

\subsection{Pruning as constrained optimization}

Optimal brain surgery relies on removing and adapting weights such that the loss is least negatively affected, thus it behooves us to write the problem as a constrained optimization problem. From the Gaussian approximation discussed in \cref{sec:taylor} obtained by quadratically expanding the log likelihood loss ${-}\log p {\approx} \frac{1}{2} \vtheta^T \mF \vtheta$, the optimal update $\Delta \vtheta {=} \hat{\vtheta} {-} \vtheta$ (and thus also $\hat{\vtheta} {=} \vtheta {+} \Delta \vtheta$) becomes the following equality constrained quadratic optimization problem \citep{hassibi1992second}:
\begin{align}
\label{eq:general-problem}
\argmin_{\Delta \vtheta} \text{  } &\frac{1}{2} \Delta \vtheta^T \mF \Delta \vtheta \\
\text{s.t.  } &\ve_k^T \Delta \vtheta + \ve_k^T \vtheta = 0,  \forall k \in \mathcal{K} \nonumber
\end{align}
where $\mF$ is positive semi-definite and $\mathcal{K}$ is the set of $K$ indices that are pruned (i.e., set to zero).

\paragraph{General solution}

We denote $\mE_K = \begin{bmatrix} \ve_{1} & \ve_{2} & \ldots & \ve_{K} \end{bmatrix}^T \in [0, 1]^{K \times P}$ as a matrix of which the row vectors are canonical basis vectors $\ve_k \in \R^P$ that select the elements to be pruned. One of the most standard approaches to solve \cref{eq:general-problem} is using Langrange multipliers, which results in a general closed-form solution for the expected increase in loss $\mathcal{L}$ and optimal weight update $\Delta \vtheta$:
\begin{align}
\label{eq:general-cost}
\mathcal{L} &= \frac{1}{2} (\mE_K \vtheta^*)^T \left(\mE_K \mF^{-1} \mE_K^T \right)^{-1} \mE_K \vtheta \\
\label{eq:general-update}
\Delta \vtheta &= -\mF^{-1} \mE_K^T \left( \mE_K \mF^{-1} \mE_K^T \right)^{-1} \mE_K \vtheta
\end{align}
which we use to derive unstructured, semi-structured, structured for modern Fisher approximations (see \cref{sec:derive-single-element,sec:derive-single-row-col,sec:derivation-multiple-structured}). The same general form of \cref{eq:general-cost,eq:general-update} appears in prior LLM pruning work \cite{kurtic2022optimal}, but only for much simpler layer-wise pruning and no structured pruning.

% %\newpage
\section{LLM Surgeon}
\label{method}
\vspace{-1.0em}
This section describes the components of our method, \textbf{LLM Surgeon}, summarised in \cref{alg:pseudocode}.
\vspace{-1.0em}

\algrenewcommand\algorithmicrequire{\textbf{Input:}}
\algrenewcommand\algorithmicensure{\textbf{Output:}}

\begin{algorithm}[t]
\caption{LLM Surgeon (\textit{structured})}\label{alg:pseudocode}
\begin{algorithmic}
\Require initial weights $\vtheta^0$, target size $\alpha$, and data $\mathcal{D}$ \\
\textbf{For} shot $t$ \text{in} [1, 2, \ldots, $T$] \\
    %\hspace{1em} \textbf{Compute:} approximate curvature $\mG_1, \mA_1$ from data $\mathcal{D}$ (\text{optionally also } $ \mG_2, \mA_2) $   \Comment{\cref{sec:estimating-curvature}} \\
    \hspace{1em} \textbf{Compute:} approximate curvature $\mG, \mA$ from data $\mathcal{D}$ \Comment{\cref{sec:estimating-curvature}} \\
    % \hspace{1em} \textbf{Compute:} costs per row/column $ \mathcal{L}_r, \mathcal{L}_c$ from $\mG_1, \mA_1, (\mG_2, \mA_2) $ \Comment{\cref{sec:compute-costs}} \\
    \hspace{1em} \textbf{Compute:} costs per row/column $ \mathcal{L}_r, \mathcal{L}_c$ from $\mG, \mA $ \Comment{\cref{sec:compute-costs}} \\
    \hspace{1em} \textbf{Compute: } threshold $\tau$ using $\mathcal{L}_r$ and $\mathcal{L}_c$ given target size $\alpha_t$ \Comment{\cref{sec:dynamic-thresholding}} \\
    \hspace{1em} \textbf{Select: } rows and columns to remove $\mE_R$, $\mE_C$ based on $\tau$  \Comment{\cref{sec:dynamic-thresholding}} \\
    %\hspace{1em} \textbf{Compute: } weight update $\Delta \vtheta^{t-1}$ based on $\mE_R, \mE_C$ and $\mG_1, \mA_1, (\mG_2, \mA_2)$ \Comment{\cref{sec:correlated-weight-updates}} \\
    \hspace{1em} \textbf{Compute: } weight update $\Delta \vtheta^{t-1}$ based on $\mE_R, \mE_C$ and $\mG, \mA$ \Comment{\cref{sec:correlated-weight-updates}} \\
    \hspace{1em} \textbf{Update:} remaining weights $\vtheta^t \leftarrow \vtheta^{t-1} + \Delta \vtheta^{t-1}$ \Comment{\cref{sec:multiple-shot}} \\
    \hspace{1em} \textbf{Optionally:} $\vtheta^t \leftarrow \text{low-rank update}(\vtheta^{t}) $ \Comment{\cref{sec:interleaved-lora}} \\
\textbf{Output:} compressed weights $\hat{\vtheta} = \vtheta^T$ 
\end{algorithmic}
\end{algorithm}

\subsection{Estimating loss landscape curvature}
\label{sec:estimating-curvature}

Accurate pruning relies on approximating the local curvature accurately while overcoming the memory cost associated with storing the true curvature. Specifically, even with the block-wise approximation of \cref{eq:block-fisher}, $\mF \in \R^{RC \times RC}$ requires summing $N$ large $RC \times RC$ matrices, too large to practically fit in memory. Instead, we adapt the KFAC approximation \citep{martens2015optimizing} that assumes independence of activations and derivatives, approximating an expectation of Kronecker products as a Kronecker product of two expectations $\E[ \vg_{l,n} \vg_{l,n}^T \otimes \va_{l,n} \va_{l,n}^T ] \approx \E[ \vg_{l,n} \vg_{l,n}^T] \otimes \E[\va_{l,n} \va_{l,n}^T ] $, allowing layer-wise Fisher blocks to be approximated as $\mF_l \approx \widetilde{\mF}_l$, where
\begin{align}
\vspace{-2em}
\widetilde{\mF_l} = \mG_l \otimes \mA_l \hspace{1em} \text{, 
 with } \mG_l = \frac{1}{\sqrt{N}} \sum\nolimits_{n=1}^N \vg_{l,n} \vg_{l, n}^T \text{   and      }  \mA_l = \frac{1}{\sqrt{N}} \sum\nolimits_{n=1}^N \va_{l,n} \va_{l, n}^T 
\vspace{-1em}
\end{align}
constructed from activations $\va_{l, n} \in \R^{C}$ from forward passes and gradients $\vg_{l, n} \in \R^{R}$ from backward passes \citep{eschenhagen2024kronecker}. The approximation originates from optimization literature, but has recently gained popularity for other problems that require curvature approximations \citep{immer2022invariance,van2023learning}, including structured pruning in \cite{wang2019eigendamage}.

An additional advantage of approximating Fisher blocks as Kronecker products is that the inverse becomes particularly easy to compute $\vspace{-0.2em}\widetilde{\mF}^{-1} = \mG^{-1} \otimes \mA^{-1}$, thus only requires inverting the factors. This fact allows us to never explicitly construct large $RC{\times}RC$ matrices in memory that make up $\widetilde{\mF}$ and $\widetilde{\mF}^{-1}$, but rather directly work with the much smaller matrices $\mG$ and $\mA$.

\subsection{Computing costs in final loss}
\label{sec:compute-costs}

The number of possible combinations in which weights can be removed grows (supra-)exponentially in parameter count, making it infeasible to estimate a separate cost $\mathcal{L}$ for each such removal. A common strategy, therefore, is to treat weights independently when computing removal costs $\mathcal{L}$. We also follow this strategy, but note that this does not necessarily imply that we have to make such same strong independence assumption for the weight updates $\Delta \vtheta$ after selecting weights to be removed. Unlike most prior work, we present correlated weight updates by taking into account off-diagonal elements of the Fisher approximation in \cref{sec:correlated-weight-updates}.

For semi-structured and unstructured we use independent costs for individual weight elements $k {\in} [1, RC]$, and for structured use independent costs for all rows $r{\in} [1, R]$ and columns $c {\in} [1, C]$. We find that we can derive the appropriate costs from the general cost formula \cref{eq:general-cost} by letting $\mE {=} \ve_k \in \R^{RC}$ where the single one-hot element at index $k$ of canonical basis vector $\ve_k$ selects the weight to remove. For structured pruning, we similarly select rows $r$ and columns $c$, by setting $\mE {=} \ve_r^T {\otimes} \mI {\in} \R^{C \times RC}$ or $\mE {=} \mI {\otimes} \ve_c {\in} \R^{R \times RC}$ with $\ve_r {\in} \R^R$, $\ve_c {\in} \R^C$. Plugging into \cref{eq:general-cost}, we find:
\begin{align}
\begin{split}
\mathcal{L}_k = \frac{1}{2} \frac{(\vtheta_k)^2}{[\mG^{-1} \otimes \mA^{-1}]_{kk}}
\end{split}, \hspace{1em}
\begin{split}
\mathcal{L}_r = \frac{1}{2} \frac{\vtheta_r^T \mA \vtheta_r}{[\mG^{-1}]_{rr}}
\end{split}, \hspace{1em}
\begin{split}
\mathcal{L}_c = \frac{1}{2} \frac{\vtheta_c^T \mG \vtheta_c}{[\mA^{-1}]_{cc}}
\end{split}
\end{align}
Full derivations can be found in \cref{sec:derive-single-element,sec:derive-single-row-col}. The costs for single elements $\mathcal{L}_k$ are equivalent to those found in optimal brain surgeon \citep{hassibi1992second} and $\mathcal{L}_r$ and $\mathcal{L}_c$ closely resemble structured brain surgeon of \citep{wang2019eigendamage}, but in our case derived for matrix rows and columns (see \cref{sec:derive-single-row-col}). Given curvature estimates, costs for either removing all weights or all rows and columns can be computed in parallel. In addition, we derive costs for the more general sum of Kronecker factor approximation $\widetilde{\mF} \approx \mG_1 \otimes \mA_1 + \mG_2 \otimes \mA_2$ in \cref{sec:extending-curvature-estimates} through an eigendecomposition.

\subsection{Dynamic weight allocation with global threshold}
\label{sec:dynamic-thresholding}

\begin{wrapfigure}{r}{0.50\textwidth}
  \vspace{-1.6em}
  \begin{center}
    \includegraphics[width=0.50\textwidth]{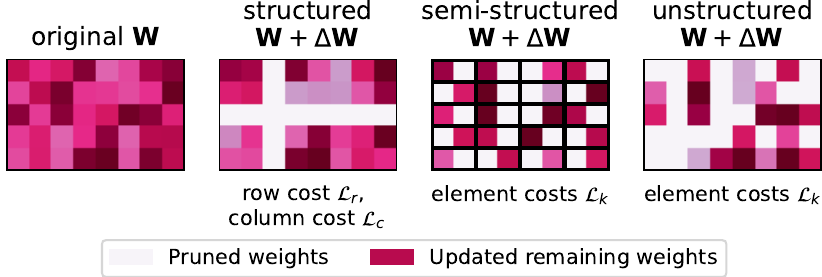}
  \end{center}
  \vspace{-1.6em}
  \caption{General framework for structured, semi-structured and unstructured compression.}
  \vspace{-0.4em}
  \label{fig:pruning-structures}
\end{wrapfigure}

Unlike prior works that compress layer-by-layer \citep{frantar2023sparsegpt}, we use a global threshold $\tau$ enabling a \textit{dynamic allocation of sparsity levels across layers}, pruning most where it hurts the least. Our method can compress a model to a specifically chosen target size $\alpha$, defined as the fraction of weights that should remain, i.e. stay non-zero after compression. In all structured, semi-structured, and unstructured pruning (\cref{fig:pruning-structures}), we select as many weights for removal so that the target size $\alpha$ is reached that inflict the least possible costs $\mathcal{L}$, as computed according to \cref{sec:compute-costs}. For unstructured pruning, this is as simple as sorting the costs for all weights $\mathcal{L}_k$ in the network and setting a global threshold $\tau$ such that $\alpha$ fraction of weights fall within the threshold $\mathcal{L}_k \leq \tau$. For M:N semi-structured pruning, we sort the M costs of each N consecutive weights and select the M weights with lowest cost. In case of a multi shot schedule (see \cref{sec:multiple-shot}) we also sum the M lowest costs in each block to find a cost per block, sort costs per block across the entire network, and similar to the unstructured case set a global threshold $\tau$ such that an $\alpha$ fraction of weights fall within threshold. Lastly for structured pruning, we perform a sorting appropriately weighted by the number of elements that make up a row or column and set the global threshold $\tau$ such that $\alpha$ fraction of all weights fall within the threshold. Then we remove all rows and columns that fall within the threshold $\mathcal{L}_r, \mathcal{L}_c \leq \tau$. % Details for all unstructured, semi-structured and structured global thresholding can be found in \cref{sec:dynamic-thresholding-details}.

\subsection{Correlated weight updates}
\label{sec:correlated-weight-updates}

Like most other pruning methods, we prune multiple weights
at once \citep{frantar2023sparsegpt,wang2019eigendamage}. To arrive at pruning costs and weight updates for pruning multiple weights, it is common to compute costs and updates for individual weights (or sets of weights) independently and add them together to arrive at a joint pruning cost. In LLM Surgeon, we argue that it's better to consider weight updates jointly instead of independently. 
After selecting the set of weights for pruning, we can often afford to compute a single correlated weight update associated to the joint removal of multiple weights, instead of naively summing weight updates associated to individual removals. 
We derive such correlated weight updates below. Note that, for the expected cost computation, we do assume that the row, column or weight costs are independent, as the number of possible combinations of weights to prune grows too large to compute within reasonable time.

\paragraph{Fast unstructured / semi-structured correlated weight updates}

Mathematically, we represent pruned weights as $\mE_K{=}\begin{bmatrix}\ve_1 & \ve_2 & {\ldots} & \ve_{R'}\end{bmatrix}^T {\in} \R^{K \times RS}$, where $\ve_r {\in} \R^{R'}$ are one-hot canonical basis vectors selecting the weights for removal. As each element $k$ has a unique associated row $r$ and column $c$ index, we can consequently also use canonical basis vectors for these respective rows $\mE_R {\in} \R^{K \times R}$ and columns $\mE_C {\in} \R^{K \times C}$ (i.e., we have $[\mE_R]_i \otimes [\mE_C]_i {=} [\mE_K]_i$ is satisfied for all $i$).

We derive unstructured weight updates in \cref{sec:derive-single-element}, by considering eigendecompositions $\mG = \mK_1 \mS_1 \mK_1^T$, $\mA = \mK_2 \mS_2 \mK_2$ of the Fisher approximation $\mF \approx \mG \otimes \mA$, which from \cref{eq:general-update} yields:

\vspace{-1.3em}
\begin{align}
\label{eq:unstructured-update}
\smash{
\Delta \mW = \mG^{-1} \Big( \mK_1 \Big( \underbrace{\widebar{\mK}_1^T \widebar{\mW}^{-1} \widebar{\mK}_2 \oslash \mS }_{K \times K\vspace{-8em}} \Big)^{-1} \mK_2 \Big) \mA^{-1}
} 
\end{align}
\vspace{-0.5em}

where $\oslash$ is element-wise division, and for brevity use bar notation $\widebar{\mK}_1 {=} \mE_K \mK_1$, $\widebar{\mK}_2 {=} \mE_K \mK_2$, $\widebar{\vtheta} {=} \mE_K \vtheta$, and $\mS {=} \text{diag}(\mS_1) \text{diag}(\mS_2)^T {\in} \R^{R \times C}$, and $\text{diag}(\cdot)$ vectorises matrix diagonals.

Programmatically, we always avoid explicitly representing large matrices $\widetilde{\mF}$ and $\widetilde{\mF}^{-1}$ in memory, but rather compute relevant quantities from their factors. Likewise, we never represent sparse matrices $\mE_K$, $\mE_R$ or $\mE_C$ in memory, but instead work with a lists of indices of the one-hot elements directly. For example, we can cheaply construct $\widebar{\mK}_1 {=} \mE_R \mK_1 \in \R^{K\times R}$ and $\widebar{\mK}_2 {=} \mE_C \mK_2 \in \R^{K \times C}$, by copying row vectors, and the vector $\widebar{\vtheta} {=} \mE_K \vtheta {=} \mE_R \mW \mE_C^T \in \R^K$ by indexing all pruned weights.

\paragraph{Maximum number of correlated weights}
The main computational bottleneck is the $K {\times} K$ matrix inverse in \cref{eq:unstructured-update}. To control compression speed, we can split pruned weights into disjoint subsets $K{=}K_1{\cup}K_2{\cup}\ldots$, such that each subset $K_i$ does not exceed the set maximum number of correlated weights $K_i {\leq} m$, and sum associated independent updates. Using less correlation by setting a lower $m$ allows trading compression quality for speed.

\paragraph{Fast structured correlated weight updates}
\label{sec:fast-structured}
Unlike the general case which requires inverting a $K \times K$ matrix for $K$ correlated weights, we find that weight updates with the Kronecker factored Fisher approximation $\tilde{\mF} = \mG \otimes \mA$ only require inverting a $R' \times R'$ matrix when removing $R'$ rows or a $C' \times C'$ matrix when removing $C'$ columns. The updates are much cheaper than we would have expected based on the effective number of weights in those rows and columns, which would imply inverting $R'C \times R'C$ or $RC' \times RC'$ matrices. In practice, this leads to a significant speed-up for structured pruning and weight updates that take into account correlations between rows or columns. When removing $R'$ rows, $r_1, r_2, \ldots r_{R'}$, or the $C'$ columns $c_1, c_2, \ldots, c_{C'}$, with $1{<}R'<R$ and $1{<}C'{<}C$, we denote one-hot vectors selecting all rows and columns to be removed respectively as $\mE_{R'} = \begin{bmatrix} \ve_1 & \ve_2 & \ldots & \ve_{R'} \end{bmatrix}^T \in \R^{R' \times R}$ and $\mE_{C'} = \begin{bmatrix} \ve_1 & \ve_2 & \ldots & \ve_{C'} \end{bmatrix}^T \in \R^{C' \times C}$. We find weight updates associated to removing the $R'$ rows by setting $\mE_K = \mE_{R'} \otimes \mI$ or $\mE_K = \mI \otimes \mE_{C'}$:
\begin{align}
\begin{split}
    \text{remove multiple $R'$ rows: } & \\
    \text{remove multiple $C'$ columns: } & 
\end{split}
\begin{split}
\Delta \mW &= -  \widebar{\mW} (\mE_{C'} \mA^{-1} \mE_{C'}^T)^{-1} (\mA^{-1} \mE_{C'}^T) \\
\Delta \mW &= -\mG^{-1} \mE_{R'}^T (\mE_{R'} \mG^{-1} \mE_{R'}^T)^{-1} \widebar{\mW}
\end{split}
\end{align}
From here, it is clear that the special case of removing a single row $r$ or column $c$ under Kronecker approximation involves inverting a $1 \times 1$ matrix, and thus only requires scalar division:
\begin{align}
\begin{split}
    \text{remove single row $r$: } \textcolor{white}{\Big|} 
\Delta \vtheta &= - \frac{\mG^{-1}\ve_r \otimes \vtheta_r}{[\mG^{-1}]_{rr}} 
\end{split}
\begin{split}
    \text{, or single column $c$: } \textcolor{white}{\Big|}  
\Delta \vtheta &= - \frac{\vtheta_c \otimes \mA^{-1} \ve_c}{[\mA^{-1}]_{cc}}
\end{split}
\end{align}
in accordance to independent structured updates in \cite{wang2019eigendamage}, for convolutional filters. We have thus extended existing structured weight updates to rows and columns, and derived update rules that also consider correlation between structured groups (in our case the rows and columns).

\subsection{Multi shot pruning schedule}
\label{sec:multiple-shot}

To improve the performance-to-sparsity ratio, we propose pruning in multiple shots. We theoretically justify this multi-shot approach by noting that the surrogate loss landscape $q$ relies on a Taylor expansion (\cref{eq:taylor}) that only holds locally and thus becomes unreliable for larger jumps $\Delta \vtheta$ in parameter space. We mitigate this by pruning in multiple $T{>}1$ shots, $t \in [1, 2, \ldots, T]$, each resulting in a smaller weight update $\Delta \vtheta$ after which the curvature of the loss surface can be re-estimated. When pruning to target size $\alpha$, ie. removing $1{-}\alpha$ of total weights, we choose a schedule $\alpha_t$ starting at $\alpha_0=1$ and ends with $\alpha_T{=}\alpha$, such that after $T$ shots, exactly $\alpha$ fraction of the total weight remain. Empirically, we find that a linear schedule for $\alpha_t$, as formulated in \cref{results}, monotonically improves pruning performance with more shots, and that higher sparsity levels typically require more shots (see \cref{sec:ablation-shots}).
Multi-shot pruning allows one to spend (linearly in $T$) more computation to improve the final compression performance. 

\subsection{Interleaved low-rank first-order corrections}
\label{sec:interleaved-lora}

We propose optional interleaved low-rank first-order corrections to further improve compression performance. So far, we assumed parameters are in a local optimum when finding a closed-form solution to the quadratic constraint problem. In practice, however, this assumption likely does not hold since (i) the neural network may not be optimised to the minimum, (ii) a different loss may be used for compression than used for training, or (iii) we prune in multiple shots (\cref{sec:multiple-shot}) inevitably causing weights to diverge from the optimum. To mitigate this, we consider first-order corrections by interleaving pruning shots with low-rank adaptations of weights $\mW_l {+} \mU \mV$ (LoRA, by \citep{hu2021lora}), commonly used in LLM finetuning. We always absorb updates after each shot, so that the next loss estimate $q$ is closer to the optimum and underlying assumptions are likely to hold more closely. By absorbing LoRA updates between shots, the sum of low-rank updates can have a higher rank than individual updates. That is, we have $\text{rank}(\mU^1\mV^1 {+} \mU^2\mV^2 {+} \ldots {+} \mU^T \mV^T) \geq \text{rank}(\mU^t\mV^t)$ for the updates $\mU^t \mV^t$ at any shot $t$, with equality only arising if updates lie exactly in the same subspace which is unlikely to ever occur in practice. This insight could also be used during regular LoRA finetuning and may therefore be useful outside the context of model compression to allow more expressive low-rank model adaptation, at negligible cost.

% \vspace{-1em}
% \subsection{Trading efficiency and performance at prune time}
% Our method allows trading improved inference time performance at the cost of more compute during prune-time, through turning several 'knobs': increasing the number of shots (linear) which improves the quality of the updates, and increasing the maximum number of correlations (cubic).%, depending on pruning type)
% By increasing the maximum number of correlations between weights, one can trade (cubically increasing) compute for better estimates of the updates. 
% Note that this scales more efficiently for structured pruning than for semi-structured or unstructured pruning.
% Likewise, multi shot pruning allows one to spend (linearly in $T$) more computation to improve the final compression performance. 
% Since a model only needs to be compressed once after which it can be used many times for deployment, being able to spend additional compute to improve final compression performance is a desirable property. 

%\newpage
% \vspace{-1.5em}
\section{Results}
% \vspace{-1em}
\label{results}

\begin{table}[t]
\caption{Structured compression of large language models on wikitext-2 data.}
% \vspace{-0.5em}
\begin{center}
    \resizebox{\linewidth}{!}{
    \begin{tabular}{l|c|cccc|cc}
    & \bf{}
    & \multicolumn{5}{c}{\bf{Test performance (PPL)}} \\
    \bf{Method} & \bf{Target size}
    & \multicolumn{1}{c}{OPT (125m)}
    & \multicolumn{1}{c}{OPT (1.3b)}
    & \multicolumn{1}{c}{OPT (2.7b)}
    & \multicolumn{1}{c}{OPT (6.7b)}
    & \multicolumn{1}{c}{Llama-v2 (7b)}
    % & \multicolumn{1}{c}{Llama-v2 (13b)}
    \\
\hline
Baseline & 100\% & \bf{27.65}
%& \bf{22.00}
& \bf{14.62} & \bf{12.47} & \bf{10.86} 
& \bf{5.12} %& \bf{4.57} 
\\
% Baseline + tuned & 100\% & \bf{23.31}
%& \bf{18.62}
% & . & & & \\
\hline
Magnitude
& 90\% & 767.2 & 894.4 & 1229 & 3464 & 36746 %& 7718 
\\
$\mI \otimes \mI$
& 80\% & 4685 & (1278) & 2788 & 16747 & 347960 %& 22463 
\\
& 70\% & 17970 & (3098) & 9255 & 17312 & 41373 %& 43555 
\\
% $\hline
% L-OBD
% & 90\% & & & & & \\
% $\text{diag}(\mI \otimes \mA)$
% & 80\% & & & & \\
% single shot
% & 70\% & 2473 & & & & \\
% % & 60\% & & 5443 & & \\
% % & 50\% & & & & \\
\hline
L-OBD
% & 90\% & 35.95 & 24.55 & 17.69 & 27.20 & 14259 \\
% $\text{diag}(\mI \otimes \mA)$
% & 80\% & 70.25 & 97.90 & 3236 & 7570 & 15630\\
% multi shot
% & 70\% & 207.2 & 1793 & 7233 & 7628 & 21386 \\
& 90\% & 33.3 & 20.76 & 17.69 & 27.20 & 14259 \\
$\text{diag}(\mI \otimes \mA)$
& 80\% & 94.14 & 1392 & 3236 & 7570 & 15630\\
multi shot
& 70\% & 545.6 & 2147 & 7233 & 7628 & 21386 \\

% \hline
% L-OBD
% & 90\% & 27.65 ? & 14.65 ? & 26.18 ? & gpu:2+3 \\
% $\text{diag}(\mI \otimes \mA)$
% & 80\% & 29.24 ? & 16.36 ? & 119.7 ? \\
% few shot ? old ?
% & 70\% & 35.42 ? & 20.09 ? & 4770 ? \\
% % & 60\% & 48.31 & 28.69 & \\
% % & 50\% & 81.48 & 50.02 & \\
\hline
K-OBD
& 90\% & 27.97 & 14.68 & 11.96 & 10.53 & 5.48  % 8.11 
\\
$\text{diag}(\mG \otimes \mA)$
& 80\% & 29.89 & 15.63 & 12.47 & 11.28 & 9.14  % 12.15 
\\
multi shot
& 70\% & 36.54 & 18.29 & 14.53 & 13.03 & 15.43 % 21.23 
\\
& 60\% & 47.54 & 24.65 & 18.09 & 16.21 & 28.03 % 39.95 
\\
& 50\% & 75.95 & 37.68 & 26.68 & 25.54 & 46.64  %94.66 
\\
\hline
LLM Surgeon (\textbf{ours})
& 90\% & 28.29 & 14.73 & 12.00 & 10.82 & 5.43 %& 
\\
$\mG \otimes \mA$
& 80\% & 29.37 & 15.27 & 12.37 & 11.22 & 7.29 %& 
\\
within row/col cor. $\Delta$
& 70\% & 32.46 & 16.60 & 13.16 & 11.83 & 10.85 %& 
\\
& 60\% & 39.82 & 19.40 & 14.79 & 12.94 & 16.67 %& 
\\
& 50\% & 51.48 & 23.81 & 18.01 & 15.38 & 25.62 %& 
\\
\hline
LLM Surgeon (\textbf{ours})
& 90\% & 28.01 & 14.70 & 12.02 & 10.77 & 5.25 %5.27 %& 
\\
$\mG \otimes \mA$
& 80\% & 28.73 & 15.12 & 12.27 & 11.02 & 6.18 %6.40 %& 
\\
full cor. $\Delta$
& 70\% & 31.82 & 16.24 & 12.92 & 11.64 & 7.83 %8.12 %& 
\\
& 60\% & 38.47 & 18.45 & 14.23 & 12.58 & 10.39 %& 
\\
& 50\% & 49.78 & 22.95 & 17.15 & 14.90 & 15.38 %& 
\\
\hline
\end{tabular}
}
\end{center}
\label{tab:results-structured}
\vspace{-1.0em}
\end{table}

We compare compression performance of LLM Surgeon on language modeling tasks on OPT \citep{zhang2022opt} and Llama-v2 \citep{touvron2023llama} model families, using data from wikitext-2 dataset (\cref{sec:datasets}). For compression, we use 128 sequences with a sequence length of 2048 tokens from the training data set and evaluate test perplexity (PPL) on the standard test split. In our experiments, we use a linear sparsity schedule $\alpha_t {=} 1{-}t ( \frac{1 - \alpha}{T} ) $ at each shot $s$ before reaching the final sparsity $\alpha$. We use 40 shots at $\alpha{=}0.5$ sparsity and report intermediate compression rates, effectively using $T{=}8$ shots for $\alpha{=}0.9$, $T{=}16$ for $\alpha{=}0.8$, $T{=}24$ for $\alpha{=}0.7$, and $T{=}32$ for $\alpha{=}0.6$. We compare against magnitude pruning, L-OBD, SparseGPT and K-OBD baselines. The K-OBD and LLM Surgeon use the multi shot procedure of \cref{sec:multiple-shot} using $T{=}40$ shots for structured pruning and $T{=}5$ shots for semistructured and unstructured pruning. Further details are found in \cref{sec:experimental-details}.

\subsection{Structured Compression}
\label{sec:results-structured}

Structured compression of rows and columns enables direct savings in memory and compute through a straight reduction of matrix dimensions in the model. 
% We consider both LLM Surgeon with weight updates that assume independence between rows and columns (`within row/col correlated updates $\Delta$') and updates correlated both within and between rows and columns (`fully corr. $\Delta$'), 
For LLM surgeon, we consider in \cref{sec:correlated-weight-updates} weight updates with different levels of correlations: limited to correlations within rows and columns, and correlations both within and between rows and columns.
We further compare against magnitude pruning, which only uses weight magnitudes, L-OBD, which only uses activations, and K-OBD, which also uses Kronecker-factored curvature but assumes full independence and thus only prunes without updating remaining weights. 
We report results in \cref{tab:results-structured}, and observe that more correlations results in better performance, with the largest improvements for the Llama-v2 model family.

While a 50\% structured compression is not better than a smaller model of similar size, LLM Surgeon allows us to reduce model size 
%by 10\% without loss and 
by up to 30\% with minimal loss, without training a smaller model from scratch \cref{fig:horizontal-overview}. 
In our structured compression experiments our proposed LLM Surgeon method outperforms all baselines and achieves the best performance for each compression target size.

\subsection{Interleaved low-rank updates}

\begin{wraptable}{h}{65mm}
\vspace{-2.2em}
\caption{Structured compression of OPT-125m on wikitext-2 using interleaved LoRA updates}
\vspace{-0.75em}
\begin{center}
\small
    \begin{tabular}{l|c c c}
& \textbf{Target} & \textbf{without} & \textbf{with} \\
& \textbf{Size} & \textbf{LoRA} & \textbf{LoRA} \\
\hline
Pretrained & 100\% & 27.65 & 23.35 \\
\hline
LLM Surgeon \hspace{0.8em}
& 90\% & 28.01 & 24.16 \\
(\textbf{ours}) 
& 80\% & 28.73 & 25.25 \\
$\mG \otimes \mA$
& 70\% & 31.82 & 28.86 \\
full cor. $\Delta$
& 60\% & 38.47 & 31.26\\
& 50\% & 49.78 & 36.50 \\
\end{tabular}
\vspace{-2em}
\end{center}
\end{wraptable}

Additionally, we assess compression performance in conjunction with the proposed first-order corrections using the interleaved low-rank adaptation described in \cref{sec:interleaved-lora}. We find that LoRA improves compression performance in the smallest 125m model, but not in larger models. We hypothesise that larger models are more prone to overfitting on the relatively few batches of wikitext-2 data used to compress the model. Nevertheless, we conclude that interleaved LoRA can be useful in cases, and recommend first using the proposed method without interleaved updates and, if enough data is available for compression, optionally using it if it improves performance.

% LLM Surgeon (\textbf{ours})
% & 100\% & 24.11/23.35 & 13.60/14.24 & \\
% $\mG \otimes \mA$
% & 90\% & 24.59/24.16 & 14.59/14.16 & \\
% full cor. $\Delta$ + interleaved LoRA
% & 80\% & 25.71/25.25 & 15.48/15.62 & \\
% & 70\% & 28.49/28.86 & & \\
% & 60\% & 31.90/31.26 & & \\
% & 50\% & 37.50/36.50 & 31.84/29.9? & \\
% \hline

\subsection{Semi-structured Compression}
\label{sec:results-semistructured}

For 2:4 semi-structured pruning, we compare LLM Surgeon with magnitude pruning, which only uses weight magnitudes, single-shot L-OBD, which only uses activations, and single-shot K-OBD, which also uses Kronecker-factored curvature but assumes full independence and thus only prunes without updating remaining weights as well as the recent state-of-the-art SparseGPT~\citep{frantar2023sparsegpt}. We report test performance after 50 \% (2:4) semi-structured compression on wikitext-2 data in \cref{tab:results-semistructured}. We empirically find that considering more weight correlations results in improved final performance after compression. Our proposed LLM Surgeon is competitive with prior work outperforming all baselines in terms of test set perplexity (PPL).

\begin{table}[h]
\vspace{-1em}
\caption{Semi-structured 2:4 compression  for large language models on wikitext-2 data.}
\vspace{-1em}
\begin{center}
    \resizebox{\linewidth}{!}{
    \begin{tabular}{lcr|c|cccc}
    & & & \bf{Target}
    & \multicolumn{4}{c}{\bf{Test performance (PPL)}} \\
    \bf{Method} & & $\mF \approx \ $ & \bf{size}
    & \multicolumn{1}{c}{OPT (125m)}
    & \multicolumn{1}{c}{OPT (1.3b)}
    & \multicolumn{1}{c}{OPT (2.7b)}
    & \multicolumn{1}{c}{OPT (6.7b)}
    % & \multicolumn{1}{c}{Llama-v2 (7b)}
    % & \multicolumn{1}{c}{Llama-v2 (13b)}
    \\
\hline
Baseline & & & 100\% & \bf{27.65} 
& \bf{14.62} & \bf{12.47} & \bf{10.86}
%& \bf{5.12} %& \bf{4.57} 
\\
\hline
Magnitude & & $\mI \otimes \mI$
& 50\% & 342.04 & 379.57 & 1106.01 & 187.29 %& 111.92 %& 8.28 
\\
L-OBD & & $\text{diag}(\mI \otimes \mA)$
& 50\% & 87.26 & 44.92 & 41.40 & 27.36 %& 11.27 %& 8.48 
\\
K-OBD & & $\text{diag}(\mG \otimes \mA)$
& 50\% & 68.74 & 27.22 & 20.23 & 15.55 %& 10.22 %& 
\\
SparseGPT & & $\mI \otimes \mA$
& 50\% & 45.51 & 29.44 & 14.92 & 13.01 %& 14.66 %& 7.08 
\\
LLM Surgeon (\textbf{ours}) & & $\mG \otimes \mA$
& 50\% & 44.64 & 25.10 & 14.64 & 12.10 %& * % & 
\\
\hline
\end{tabular}
}
\end{center}
\label{tab:results-semistructured}
\end{table}

\vspace{-1em}
\subsection{Unstructured Compression}
\label{sec:results-unstructured}

For unstructured pruning, we repeat the same experiments as structured pruning case described in \cref{sec:results-structured}. In \cref{tab:results-unstructured}, we report final test performance in terms of perplexity (PPL) on wikitext-2 after compressing LLMs of different sizes of OPT and Llama-v2 family. Overall, we find that methods with more accurate approximations of the curvature landscape and that account for more correlations perform better. The proposed LLM Surgeon outperforms all baselines, reaching the highest test performance across target sizes.

\begin{table}[h!!]
\vspace{-1em}
\caption{Unstructured compression of large language models on wikitext-2 data.}
\vspace{-1em}
\begin{center}
    \resizebox{\linewidth}{!}{
    \begin{tabular}{l|c|cccc|c}
    & \bf{Target}
    & \multicolumn{5}{c}{\bf{Test performance (PPL)}} \\
    \bf{Method} & \bf{size}
    & \multicolumn{1}{c}{OPT (125m)}
    & \multicolumn{1}{c}{OPT (1.3b)}
    & \multicolumn{1}{c}{OPT (2.7b)}
    & \multicolumn{1}{c}{OPT (6.7b)}
    & \multicolumn{1}{c}{Llama-v2 (7b)}
    % & \multicolumn{1}{c}{Llama-v2 (13b)}
    \\
\hline
Baseline & 100\% & \bf{27.65}
& \bf{14.62} & \bf{12.47} & \bf{10.86} & \bf{5.12} %& \bf{4.57} 
\\
\hline
Magnitude           & 90\% & 27.62 & 14.69 & 12.60 & 10.88 & 5.18 % & 4.59 
\\
$ \mI \otimes \mI$  & 80\% & 28.53 & 15.68 & 13.18 & 11.26 & 5.37 %& 4.67 
\\
                    & 70\% & 52.88 & 140.2 & 15.22 & 12.22 & 6.03 %& 4.87 
                    \\
% $& 60\% & 249.6 & & 23.74 & 31.42 \\
% $& 50\% & 725.6 & & 276.2 & 436 \\
\hline
L-OBD
& 90\% & 29.70 & 16.24 & 14.44 & 13.43 & 6.09\\
$\text{diag}(\mI \otimes \mA)$
& 80\% & 32.18 & 21.92 & 23.35 & 39.85 & 116.2\\
single shot
& 70\% & 49.08 & 204.7 & 274.8 & 810.4 & 6549\\
% & 60\% & 281.9 & 5443 & 4100 & \\
% & 50\% & 2391 & 10266 & 13039 & 10703 \\

% \hline
% L-OBD
% & 90\% & 33.81 & & & & \\
% $\text{diag}(\mI \otimes \mA)$
% & 80\% & 41.42 & & & \\
% multi shot
% & 70\% & 60.62 & & 74.8 & \\
% % & 60\% & 177.4 & & & \\
% % & 50\% & 1628 & & & \\
\hline
K-OBD
& 90\% & 27.64 & 14.62 & 12.09 & 36.89 & 5.13 \\
$ \mG \otimes \mA$ 
& 80\% & 27.62 & 14.37 & 130220 & 39928 & 5.19 \\
single shot
& 70\% & 27.92 & 220.1 & 23097 & 19506 & 5.60 \\
& 60\% & 29.24 & 13783 & 10331 & 33896 & 9.20 \\
& 50\% & 34.43 & 7311 & 10495 & 91506 & 118.6 \\
\hline
SparseGPT
& 90\% & 27.93 & 14.69 & 12.00 & 10.86 & 5.49 %& 4.91 
\\
$ \mI \otimes \mA $
& 80\% & 28.18 & 15.07 & 12.05 & 10.86 & 5.58 %& 4.97 
\\
& 70\% & 28.93 & 22.77 & 12.17 & 10.89 & 5.71 %& 5.07 
\\
& 60\% & 30.20 & 25.07 & 12.37 & 10.98 & 5.94 %& 5.26 
\\
& 50\% & 33.17 & 26.77 & 12.88 & 11.92 & 6.51 %& 5.70 
\\
\hline
LLM Surgeon (\textbf{ours})
& 90\% & 27.69 & 14.62 & 12.01 & 10.86 & 5.13 \\
$\mG_1 \otimes \mA_1 $
& 80\% & 27.83 & 14.66 & 12.14 & 10.87 & 5.20 \\
full cor. $\Delta$
& 70\% & 28.35 & 14.81 & 12.25 & 10.82 & 5.36 \\
multi shot 
& 60\% & 28.98 & 14.91 & 12.28 & 10.83  & 5.66 \\
& 50\% & 30.30 & 15.47 & 12.68 & 10.97  & 6.08 \\
% LLM Surgeon (\textbf{ours})
% & 90\% & 27.69 (5) & 14.62 (5) & 12.01 (5), 12.01 (40) & /10.86 (5) & 5.13 (5) \\
% $\mG_1 \otimes \mA_1 $
% & 80\% & 27.83 (5) & 14.66 (5) & 12.14 (5), 12.02 (40) & /10.87 (5) & 5.20 (5) \\
% full cor. $\Delta$
% & 70\% & 28.35 (5) & 14.81 (5) & 12.25 (5), 12.09 (40) & /10.82 (5) & 5.36 (5) \\
% multi shot 
% & 60\% & 28.98 (5) & 14.91 (5) & 12.3 (5), 12.16 (40) & /10.83 (5) &  \\
% & 50\% & 30.30 (5) & 15.47 (5) & DIED & /10.97 (5)  &  \\
\hline
\end{tabular}
}
\end{center}
\label{tab:results-unstructured}
\vspace{-1em}
\end{table}

% \subsection{Compression scaling laws}
% \label{sec:results-laws}

% It has been observed that the performance of large language models (LLMs) as a function of model size can be captured in so called scaling laws (cite?), which allows prediction of model performance of larger sizes. Similarly, we measure the performance of LLMs as a function of both model size and compression ratio and find \textit{compression scaling laws} that allow predict performance for larger model sizes and compression ratios.

\newpage
\subsection{Learned sparsity structure}
\label{sec:results-structure}

\begin{figure}[h]
    \label{fig:structure}
    \vspace{-.5em}
    \resizebox{\linewidth}{!}{
    \includegraphics[]{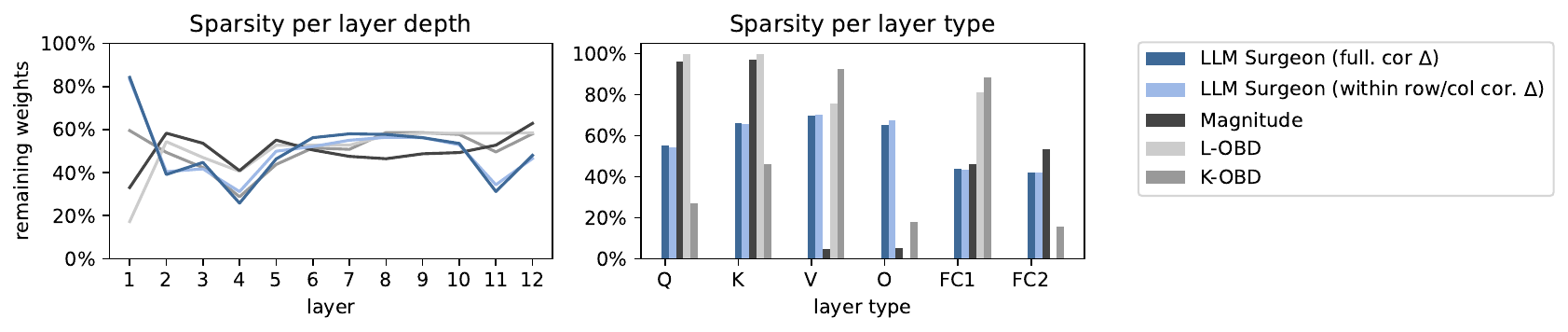}
    }
    \vspace{-2.0em}
    \caption{
    Sparsity levels obtained with structured pruning on OPT-125m by layer depth and type.
    % Sparsity levels obtained with structured pruning on OPT-125m per layer depth and per layer type. The method prunes relatively few weights in the first layer and on average prunes more in fully-connected \texttt{FC1}, \texttt{FC2} layers compared to the \texttt{Q}, \texttt{K}, \texttt{V}, \texttt{O} weights of the attention mechanism.
    } 
    \vspace{-1.0em}
\end{figure}

The proposed method can dynamically allocate sparsity across layers through global thresholds described in \cref{sec:dynamic-thresholding}. In Fig. \ref{fig:structure}, we compare total allocated sparsity levels per layer depth and per layer type after compressing a pretrained OPT-125m model. We find that the LLM Surgeon prunes relatively more in the first layer and less in middle layers. Further, we observe that a larger portions of weights are removed in fully-connected compared to attention blocks, but deviations are less compared to other methods. Dynamic allocation allows for most pruning where it hurts least.

\section{Conclusion}
\label{conclusion}

In this work, we have introduced the \textbf{LLM Surgeon} algorithm for unstructured, semi-structured and structured compression of neural networks. The work builds upon classic neural network compression approaches originating from the early 1990's that aim to find optimal pruning by expanding the curvature of the loss landscape. The method utilises modern Fisher approximations to scale accurate pruning to the realm of large language models (LLMs) with billions of parameters, while remaining practical in both memory and compute. Unlike most prior work on data-based LLM compression, we not only use weight magnitude and activations from forward passes, but also use gradient information from backward passes to relate weight removal costs to the true final objective. We improve upon prior work through more accurate approximations to the loss landscape curvature and considering more weight correlations to update remaining weights. Increasing the number of correlations and using multiple shots allows us trading off additional compute for better accuracy. Lastly, LLM Surgeon gives the first practically usable results for structured pruning of LLMs and achieves state-of-the-art results in unstructured and semi-structured large language model pruning.
% Typically, model compression only needs to happen once in practice, after which it can be deployed many times at the achieved post-compression performance. This motivates our method which, compared to several baselines, takes more time to compress but achieves the most favorable performance/compression trade-off.


\bibliography{iclr2024_conference}
\bibliographystyle{iclr2024_conference}

\appendix

\newpage
\section{Derivations for pruning}
\label{sec:structured-pruning-derivations}

%To allow for easy implementation, we derive explicit formula's for expected loss $\mathcal{L}$ and update rules in terms of both vectorised parameter vectors $\vtheta$ or weight matrices $\mW$, that relate through:
%\begin{align}
%\vtheta &= \text{vec}(\mW) \in \R^{RW} \\
%\mW &= \text{mat}(\vtheta) \in \R^{R\times W}
%\end{align}
% For updates, we use a one-hot masking vector indicating the element to be removed:
% \begin{align}
% \vq &= \text{vec}(\mQ) = \begin{bmatrix} 0 & 0 & 1 & 0 \ldots 0 \end{bmatrix}^T \in \R^{RW} \\
% \mQ &= \text{mat}(\vq) \in \R^{R \times W} 
% \end{align}
% where the 1's correspond to $q$'th index to be removed, and 0's for the other remaining indices.
% 
% The weight matrices formula's are written using element-wise/Hadamard product and $\oslash$ for element-wise division operations to more easily be implemented in a parallelisable manner. In addition, we slightly deviate from common practice by using the row-wise vectorisation operation $\text{row} : \R^{R \times C} \to \R^{RW}$ (rather than column-wise!), and denote its inverse as $\text{mat} : \R^{RW} \to \R^{R \times C}$.

Given that we use a Gaussian approximation of our loss $p {\approx} q = \mathcal{N}$ through a quadratic approximation of our log likelihood $-\log p {\approx} \frac{1}{2} (\vtheta^*)^T \mF \vtheta^*$, the most optimal compression becomes the solution to the following constrained optimization problem:
\begin{align}
\label{eq:general-problem-app}
\argmin_{\Delta \vtheta^*} \text{  } &\frac{1}{2} \Delta (\vtheta^*)^T \mF \Delta \vtheta^* \\
\text{s.t.  } &\ve_k^T \Delta \vtheta^* + \ve_k^T \vtheta^* = 0,  \forall k \in Q \nonumber
\end{align}
where $\mathcal{Q}$ is the set of $Q$ indices that are pruned.

\subsection{General solution}
Following \citep{kurtic2022optimal}, we denote pruned elements as $\mE_K = \begin{bmatrix} \ve_{q_1} & \ve_{q_2} & \ldots \end{bmatrix}^T \in [0, 1]^{|Q| \times P}$ and use the fact that solving \cref{eq:general-problem} through use of Langrange multipliers gives the general closed-form solution for cost $\mathcal{L}$ and weight update $\Delta \vtheta$:
\begin{align}
\label{eq:general-cost-app}
\mathcal{L} &= \frac{1}{2} (\mE_K \vtheta^*)^T \left(\mE_K \mF^{-1} \mE_K^T \right)^{-1} \mE_K \vtheta^* \\
\Delta \vtheta^* &= \mF^{-1} \mE_K^T \left( \mE_K \mF^{-1} \mE_K^T \right)^{-1} \mE_K \vtheta^*
\end{align}

\subsection{Removing a single element}
\label{sec:derive-single-element}

\paragraph{Optimal brain surgeon (OBS)}

To remove a single element with index $q$, we simply set $\mE_K = \ve_k^T$:
\begin{align}
\begin{split}
\mathcal{L}
&= \frac{1}{2} (\mE_K \vtheta^*)^T \left(\mE_K \mF^{-1} \mE_K^T \right)^{-1} \mE_K \vtheta \\
&= \frac{1}{2} \vtheta_k^T \frac{1}{[\mF^{-1}]_{kk}} \vtheta_k \\
&= \frac{1}{2} \frac{(\vtheta_k)^2}{[\mF^{-1}]_{kk}} \\
\end{split}\text{,  }
\begin{split}
\Delta \vtheta
&= -\mF^{-1} \mE_K^T \left( \mE_K \mF^{-1} \mE_K^T \right)^{-1} \mE_K \vtheta \\
&= -\mF^{-1} \ve_k \left( \ve_k^T \mF^{-1} \ve_k \right)^{-1} \ve^T_K \vtheta \\
&= -\frac{\vtheta_k}{[\mF^{-1}]_{kk}} \mF^{-1} \ve_k \\
\end{split}
\end{align}
which exactly correspond to the loss and updates of \textit{optimal brain surgeon} \citep{hassibi1992second}.

\paragraph{Optimal brain damage (OBD)}

We may also consider that elements are independent and the Fisher is diagonal. After noting that this implies that diagonal elements of the inverse Fisher are scalar inverses of elements in the Fisher $[\mF^{-1}]_{kk} = \frac{1}{[\mF]_{kk}}$, the formula's simplify to:
\begin{align}
\begin{split}
\mathcal{L} &= [\mF]_{kk} (\vtheta_k)^2
\end{split}\text{,  }
\begin{split}
\Delta \vtheta
&= -\vtheta_k \ve_k
\end{split}
\end{align}
which exactly corresponds to loss and updates of \textit{optimal brain damage} \citep{lecun1989optimal}.

\subsubsection*{Vectorised}

For implementation purposes, it might be convenient to have a vectorised notation $\mathcal{L}_\vtheta \in \R^{RC}$ or $\mathcal{L}_{\mW} \in \R^{R \times C}$ to calculate all expected losses in parallel:

% \paragraph{Optimal brain surgeon (OBD)}

\begin{align}
\begin{split}
\text{For OBD:  } \textcolor{white}{\Big|} & \\
\text{For OBS:  } \textcolor{white}{\Big|} &
\end{split}
\begin{split}
\mathcal{L}_{\vtheta} &= \frac{1}{2} \vtheta^* \odot \vtheta^* \odot \text{diag}(\mF) \\
\mathcal{L}_{\vtheta} &= \frac{1}{2} \vtheta^* \odot \vtheta^* \oslash \text{diag}(\mF^{-1}) 
\end{split}, \hspace{2em}
\begin{split}
\mathcal{L}_{\mW} &= \frac{1}{2} \mW^* \odot \mW^* \odot \text{mat}(\text{diag}(\mF)) \\ 
\mathcal{L}_{\mW} &= \frac{1}{2} \mW^* \odot \mW^* \oslash \text{mat}(\text{diag}(\mF^{-1})) 
\end{split}
\end{align}

% \paragraph{Optimal brain surgeon (OBS)}

% %\newpage
% \subsection{Structured pruning of rows and columns (naive / independent)}
% 
% We now consider removing an entire row or column $r \in [0, R]$ or column $c \in [0, C]$ giving a vector containing the expected losses for removing individual rows $\mathcal{L}_{r} \in \R^R$ and columns $\mathcal{L}_{c} \in \R^C$. For updates $\Delta \vtheta$ or $\Delta \mW$, we denote vectors $\vq_r \in [0, 1]^R $ and $\vq_c \in [0, 1]^C$ selecting rows and columns that are pruned with a $1$, and $0$ for those that remain.
% 
% The most naive thing to do is to assume all rows and columns are independent, but also all elements within rows and columns are independent. In this case, we could simply use the formula's from ... by summing over the expected losses and updates:
% \begin{align}
% \begin{split}
% \mathcal{L}_r &= \sum_{c=1}^C \mathcal{L}_{rc} 
% \end{split}, 
% \begin{split}
% \mathcal{L}_c &= \sum_{r=1}^R \mathcal{L}_{rc}
% \end{split}
% \end{align}
% 
% In practice, not taking into account any correlations will give bad results (cite?).

%\newpage
\subsection{Removing a single row or column}
\label{sec:derive-single-row-col}

\paragraph{Structured OBS}

If we consider the approximation $\mF \approx \mG \otimes \mA$ with known inverse $(\mG \otimes \mA)^{-1} = \mG^{-1} \otimes \mA^{-1}$, then to remove a row at index $r \in [0, R]$, we must take into account correlations within elements of that row. That is, we write matrix $\mE_K = (\ve^T_r \otimes \mI)$ containing one-hot row-vectors for all elements in row $r$. Plugging into the general solution \cref{eq:general-cost}, we find:
\begin{align}
\label{eq:single-row-loss}
\mathcal{L}
&= \frac{1}{2} \mE_K \vtheta^T \left(\mE_K \mF^{-1} \mE_K^T \right)^{-1} \mE_K \vtheta^* \nonumber \\
&= \frac{1}{2} ((\ve_r^T \otimes \mI) \vtheta^*)^T \left((\ve_r^T \otimes \mI) (\mG \otimes \mA)^{-1} (\ve_r^T \otimes \mI)^T \right)^{-1} (\ve_r^T \otimes \mI) \vtheta^* \nonumber \\
&= \frac{1}{2} \vtheta_r^T  \left(\ve_r^T \mG^{-1} \ve_r \otimes \mI \mA^{-1} \mI \right)^{-1} \vtheta_r \nonumber \\
&= \frac{1}{2} \vtheta^T (\ve_r^T \otimes \mI)  \left(\left[[\mG^{-1}]_{rr}\right] \otimes \mA^{-1} \right)^{-1} (\ve_r \otimes \mI) \vtheta_r \nonumber \\
&= \frac{1}{2} \frac{\vtheta_r^T \mA \vtheta_r}{[\mG^{-1}]_{rr}}
\end{align}
where we write $\vtheta_r = \ve_r^T \mW^* {\in} \R^C$ for the $r$'th row-vector in $\mW$. Similarly, we obtain the associated weight update:
\begin{align}
\label{eq:single-row-update}
\Delta \vtheta
&= -\mF^{-1} \mE_K^T \left( \mE_K \mF^{-1} \mE_K^T \right)^{-1} \mE_K \vtheta^* \nonumber \\
&= -\left(\mG \otimes \mA\right)^{-1} (\ve_r^T \otimes \mI)^T \left( (\ve_r^T \otimes \mI) \left(\mG \otimes \mA\right)^{-1} (\ve_r^T \otimes \mI)^T \right)^{-1} (\ve_r^T \otimes \mI) \vtheta^* \nonumber \\
&= -\left(\mG^{-1} \otimes \mA^{-1}\right)(\ve_r \otimes \mI) \left( \ve_r^T \mG^{-1} \ve_r \otimes \mA^{-1} \right)^{-1} \vtheta_r \nonumber \\
&= - \frac{1}{[\mG^{-1}]_{rr}} \left( \mG^{-1} \ve_r \otimes \mA^{-1} \mI \mA^{-1} \mI \right) \vtheta_r \nonumber \\
&= - \frac{\mG^{-1} \ve_r \otimes \vtheta_r}{[\mG^{-1}]_{rr}} 
\end{align}
arriving at a similar structured pruning update as derived in \citep{wang2019eigendamage} for convolutional filters. We can equivalently derive expected loss and update for columns, by considering $\mE_K = (\mI \otimes \ve_c^T)$. If we do so, we find the structured updates for a row $r$ or column $c$:
\begin{align}
\begin{split}
    \text{Remove row $r$:  } \textcolor{white}{\Big|} & \\
    \text{Remove column $c$: } \textcolor{white}{\Big|} & 
\end{split}
\begin{split}
\mathcal{L}
&= \frac{1}{2} \frac{\vtheta_r^T \mA \vtheta_r}{[\mG^{-1}]_{rr}} \\
\mathcal{L}
&= \frac{1}{2} \frac{\vtheta_c^T \mG \vtheta_c}{[\mA^{-1}]_{cc}}
\end{split}\hspace{2em}
\begin{split}
\Delta \vtheta
&= - \frac{\mG^{-1} \ve_r \otimes \vtheta_r}{[\mG^{-1}]_{rr}} \\
\Delta \vtheta
&= - \frac{\vtheta_c \otimes \mA^{-1} \ve_c}{[\mA^{-1}]_{cc}} \\
\end{split}
\end{align}

\paragraph{Structured OBD}

We may also assume that, when removing a row $r$, the individual elements within the row are also independent which would imply $[\mA]_{ii} = \frac{1}{[\mA^{-1}]_{ii}}$. Similarly, $[\mG]_{ii} = \frac{1}{[\mG^{-1}]_{ii}}$ when removing a column $c$. Consequently, we can simplify to:
\begin{align}
\begin{split}
    \text{Remove row $r$:  } \textcolor{white}{\Big|} & \\
    \text{Remove column $c$: } \textcolor{white}{\Big|} & 
\end{split}
\begin{split}
\mathcal{L}
&= \frac{1}{2} \mG_{rr} \vtheta_r^T \mA \vtheta_r \\
\mathcal{L}
&= \frac{1}{2} \mA_{cc} \vtheta_c^T \mG \vtheta_c
\end{split}\hspace{2em}
\begin{split}
\Delta \vtheta
&= \textcolor{white}{\Big|} -\ve_r \vtheta_r^T \\
\Delta \vtheta
&= \textcolor{white}{\Big|} -\vtheta_c \ve_c^T \\
\end{split}
\end{align}
similar form to structured OBD losses and updates as derived in \citep{wang2019eigendamage} for convolutional filters. The derivations slightly differ in that we start from the general solution \cref{eq:general-update}, circumventing the need to rederive a Langrange multipliers for each possible structure.

%\newpage
\subsection{Pruning multiple (correlated) rows and columns}
\label{sec:derivation-multiple-structured}

Let us consider the removal of $R'$ rows $r_1, r_2, \ldots r_R'$ rows or $C'$ columns with indices $c_1, c_2, \ldots, c_{C'}$, with $1{<}R'<R$ and $1{<}C'{<}C$. We denote matrices containing one-hot vectors selecting all rows and columns to be removed respectively as:
\begin{align}
\begin{split}
\mE_{R'} &= \begin{bmatrix} \ve_1 & \ve_2 & \ldots & \ve_{R'} \end{bmatrix}^T \in \R^{R' \times R}
\end{split}
\begin{split}
\mE_{C'} &= \begin{bmatrix} \ve_1 & \ve_2 & \ldots & \ve_{C'} \end{bmatrix}^T \in \R^{C' \times C}
\end{split}
\end{align}
Then, the matrix $\mE_K$ containing one-hot row vectors selecting all elements to be removed can be written as:
\begin{align}
\begin{split}
\text{Multiple rows:  } & \\
\text{Multiple columns:  } & \\
\end{split} 
\begin{split}
\mE_K = (\mE_{R'} \otimes \mI_C)
\in \R^{Q \times RC} \text{   , (with } Q = R' C \text{)} \\
\mE_K = (\mI_R \otimes \mE_{C'})
\in \R^{Q \times RC} \text{   , (with } Q = R C' \text{)}
\end{split}
\end{align}
To simultaneously remove rows and columns, we can stack the matrices with duplicate row vectors removed:
\begin{align}
\begin{split}
\text{Multiple rows and columns:} &
\end{split}
\begin{split}
\mE_K
\begin{bmatrix}
\mE_{R'} \otimes \mI_C \\
\mI_R \otimes \mE_{C'}
\end{bmatrix}
\in \R^{Q \times RC} \text{  with duplicate rows removed}
\end{split}
\end{align}
The removal of duplicate rows is required due to the few $R'C'$ overlapping elements between rows and columns, after which the total number of rows thus becomes $Q = R'C + C'R - R'C'$. We used appropriately sized identity matrices $\mI_R \in \R^{R \times R}$ and $\mI_C \in \R^{C \times C}$. For brevity, we write the vector or matrix of pruned weights $\widebar{\vtheta} := \mE_K \vtheta \in \R^{Q}$.

First, we derive the removal for $R'$ rows by defining removal matrix as $\mE_K = \mE_{R'} \otimes \mI$ and define $\widebar{\mW} := \mE_{R'} \mW \in \R^{R' \times C}$. The complete weight update for the removal of multiple rows becomes:
\begin{align}
\Delta \vtheta
&= -\mF^{-1} \mE_K^T \left( \mE_K \mF^{-1} \mE_K^T \right)^{-1} \mE_K \vtheta^* \nonumber \\
&= -(\mG \otimes \mA)^{-1} (\mE_{R'} \otimes \mI)^T \left( (\mE_{R'} \otimes \mI) (\mG \otimes \mA)^{-1} (\mE_{R'} \otimes \mI)^T \right)^{-1} (\mE_{R'} \otimes \mI) \vtheta^* \nonumber \\
&= -(\mG^{-1} \mE_{R'}^T \otimes \mA^{-1} )\left( \mE_{R'} \mG^{-1} \mE_{R'}^T \otimes \mA^{-1} \right)^{-1} \widebar{\vtheta^*} \nonumber \\
&= -(\mG^{-1} \mE_{R'}^T \otimes \mA^{-1} )\left( (\mE_{R'} \mG^{-1} \mE_{R'}^T)^{-1} \otimes \mA \right) \widebar{\vtheta^*} \nonumber \\
\Delta \mW &= -\mG^{-1} \mE_{R'}^T \left( (\mE_{R'} \mG^{-1} \mE_{R'}^T)^{-1} \widebar{\mW} \mA \right) \mA^{-1} \nonumber \\
&= -\mG^{-1} \mE_{R'}^T (\mE_{R'} \mG^{-1} \mE_{R'}^T)^{-1} \widebar{\mW}
\end{align}
Similarly, we derive the removal of $C'$ columns by defining removal matrix as $\mE_K = \mI \otimes \mE_{C'}$ and define $\widebar{\mW} := \mE_{C'} \mW \in \R^{R \times C'}$. The complete weight update for multiple column removal becomes:
\begin{align}
\Delta \vtheta
&= -\mF^{-1} \mE_K^T \left( \mE_K \mF^{-1} \mE_K^T \right)^{-1} \mE_K \vtheta^* \nonumber \\
&= -(\mG \otimes \mA)^{-1} (\mI \otimes \mE_{C'}))^T \left( (\mI \otimes \mE_{C'}) (\mG \otimes \mA)^{-1} (\mI \otimes \mE_{C'})^T \right)^{-1} (\mI \otimes \mE_{C'}) \vtheta^* \nonumber \\
&= -(\mG \otimes \mA)^{-1} (\mI \otimes \mE_{C'}))^T \left( (\mI \otimes \mE_{C'}) (\mG \otimes \mA)^{-1} (\mI \otimes \mE_{C'})^T \right)^{-1} (\mI \otimes \mE_{C'}) \vtheta^* \nonumber \\
&= -(\mG^{-1} \otimes \mA^{-1} \mE_{C'}^T ) \left( \mG \otimes \mE_{C'} \mA^{-1} \mE_{C'}^T 
 \right)^{-1} \widebar{\vtheta} \nonumber \\
\Delta \mW &= -  \mG^{-1} \mG \widebar{\mW} (\mE_{C'} \mA^{-1} \mE_{C'}^T)^{-1} (\mA^{-1} \mE_{C'}^T) \nonumber \\
&= -  \widebar{\mW} (\mE_{C'} \mA^{-1} \mE_{C'}^T)^{-1} (\mA^{-1} \mE_{C'}^T) 
\end{align}

\newpage

\section{Experimental details.}
\label{sec:experimental-details}

Code is available at: \href{https://github.com/Qualcomm-AI-research/llm-surgeon}{https://github.com/Qualcomm-AI-research/llm-surgeon}.

\subsection{Models}
\label{sec:models}

\paragraph{OPT models}

From the OPT model family (\citep{zhang2022opt}), we consider models with the following number of parameters: 125 million (125m), 1.3 billion (1.3b), 2.7 billion (2.7b), 6.7 billion (6.7b) models. We omit 350 million model due to different layer norm. We obtain the standard pre-trained checkpoints using Huggingface \citep{wolf2019huggingface} and use this as a baseline and initialisation for compression.

\paragraph{Llama-v2 models}

From the Llama-v2 model family (\citep{touvron2023llama}), we consider a model with 7 billion (7b) parameters and a model with 13 billion (13b) parameters. We obtain the standard pre-trained checkpoints using Huggingface \citep{wolf2019huggingface} and use this as a baseline and initialisation for compression.

\subsection{Datasets}
\label{sec:datasets}

\paragraph{English / Wikitext-2} The majority of the results are obtained on the Wikitext-2 dataset containing parsed subsets of the English Wikipedia \citep{merity2016pointer,wikipedia2004wikipedia}, using the default training and test sets. For fitting, we use 128 batches of 2048 characters and for testing we use the standard test set containing 4358 characters.

\paragraph{French / Wikipedia} For French data experiments, we use a subset of French wikipedia \citep{wikipedia2004wikipedia}. For fitting, we use 128 batches of 2048 characters and for testing we use a randomly selected test set containing 1067888 characters.

\paragraph{German / Wikipedia} For the Italian data experiments, we use a subset of the German wikipedia \citep{wikipedia2004wikipedia}. For fitting, we use 128 batches of 2048 characters and for testing we use a randomly selected test set containing 1112372 characters.

\paragraph{Italian / Wikipedia} For the Italian data experiments, we use a subset of the Italian wikipedia \citep{wikipedia2004wikipedia}. For fitting, we use 128 batches of 2048 characters and for testing we use a randomly selected test set containing 633177 characters.

%\paragraph{Python / CodeSearchNet} For Python data experiments, we use a subset of the CodeSearchNet corpus \citep{husain2019codesearchnet}. For fitting, we use 128 batches of 2048 characters and for testing we use a randomly selected test set containing 917601 characters.

\subsection{Mask equivalence}
\label{sec:mask-equivalence}

When comparing the equivalence of obtained pruning masks between two models $\vtheta_A$ and $\vtheta_B$ obtained by two compression methods $A$ and $B$. We always consider the case of 50\% pruning, and define the mask equivalence as the fraction of same weights that are set two zero in both models:
\begin{align}
\text{mask equivalence} = \sum_{i=1}^P \frac{\mathbf{1}([\vtheta_A]_i = 0 \text{ and } [\vtheta_B]_i = 0)}{P}.
\end{align}
where $\mathbf{1}$ denotes an indicator function that returns 1 if both weights $[\vtheta_A]_i$ and $[\vtheta_B]_i$ are zero, and returns 0 otherwise.

\subsection{SparseGPT and evaluation of baselines}

For the SparseGPT baseline, we used the official code SparseGPT code repository \citep{frantar2023sparsegpt} which allows for training and evaluation on wikitext-2. The obtained results may differ from those reported in the original paper as the C4 dataset was used there.

In this work, models were trained with the same 128 batches of the wikitext-2 training set as available in the SparseGPT codebase and are evaluated on the wikitext-2 test set using the exact same evaluation procedure.

%\newpage
\section{Technical details}

\subsection{Pseudocodes}
\label{sec:pseudocodes}

\algrenewcommand\algorithmicrequire{\textbf{Input:}}
\algrenewcommand\algorithmicensure{\textbf{Output:}}

\begin{algorithm}[H]
\caption{LLM Surgeon (\textit{structured})}\label{alg:pseudocode-structured}
\begin{algorithmic}
\Require target size $\alpha$ 
\Require initial weights $\vtheta^0$ \\
\textbf{For} shot $t$ \text{in} [1, 2, \ldots, $T$] \\
    \hspace{1em} \textbf{Compute:} approximate curvature $\mG_1, 
\mA_1$ from data (\text{optionally also } $ \mG_2, \mA_2) $   \Comment{\cref{sec:estimating-curvature}} \\
    \hspace{1em} \textbf{Compute:} costs per row/column $ \mathcal{L}_r, \mathcal{L}_c$ from $\mG_1, \mA_1, (\mG_2, \mA_2) $ \Comment{\cref{sec:compute-costs}} \\
    \hspace{1em} \textbf{Compute: } threshold $\tau$ using $\mathcal{L}_r$ and $\mathcal{L}_c$ given target size $\alpha$ \Comment{\cref{sec:dynamic-thresholding}} \\
    \hspace{1em} \textbf{Select: } rows and columns to remove $\mE_R$, $\mE_C$ based on $\tau$   \Comment{\cref{sec:dynamic-thresholding}} \\
    \hspace{1em} \textbf{Compute: } weight update $\Delta \vtheta^{t-1}$ based on $\mE_R, \mE_C$ and $\mG_1, \mA_1, (\mG_2, \mA_2)$ \Comment{\cref{sec:correlated-weight-updates}} \\
    \hspace{1em} \textbf{Update:} remaining weights $\vtheta^t \leftarrow \vtheta^{t-1} + \Delta \vtheta^{t-1}$ \Comment{\cref{sec:multiple-shot}} \\
    \hspace{1em} \textbf{Optionally:} $\vtheta^t \leftarrow \text{low-rank update}(\vtheta^{t}) $ \\
\textbf{Output:} compressed weights $\hat{\vtheta} = \vtheta^T$ 
\end{algorithmic}
\end{algorithm}

\begin{algorithm}[H]
\caption{LLM Surgeon (\textit{semi-structured / unstructured})}\label{alg:pseudocode-unstructured}
\begin{algorithmic}
\Require target size $\alpha$ 
\Require initial weights $\vtheta^0$ \\
\textbf{For} shot $t$ \text{in} [1, 2, \ldots, $T$] \\
    \hspace{1em} \textbf{Compute:} approximate curvature $\mG_1, 
\mA_1$ from data (\text{optionally also } $ \mG_2, \mA_2) $   \Comment{\cref{sec:estimating-curvature}} \\
    \hspace{1em} \textbf{Compute:} costs per element $ \mathcal{L}_k$ from $\mG_1, \mA_1, (\mG_2, \mA_2) $ \Comment{\cref{sec:compute-costs}} \\
    \hspace{1em} \textbf{Compute: } threshold $\tau$ from $\mathcal{L}_k$ and target size $\alpha_t$ (unstructured/semistructured) \Comment{\cref{sec:dynamic-thresholding}} \\
    \hspace{1em} \textbf{Select: } elements to remove $\mE_K$ based on $\tau$ (unstructured/semistructured)  \Comment{\cref{sec:dynamic-thresholding}} \\
    \hspace{1em} \textbf{Compute: } weight update $\Delta \vtheta^{t-1}$ based on $\mE_K$ and $\mG_1, \mA_1, (\mG_2, \mA_2)$ \Comment{\cref{sec:correlated-weight-updates}} \\
    \hspace{1em} \textbf{Update:} remaining weights $\vtheta^t \leftarrow \vtheta^{t-1} + \Delta \vtheta^{t-1}$ \Comment{\cref{sec:multiple-shot}} \\
    \hspace{1em} \textbf{Optionally:} $\vtheta^t \leftarrow \text{low-rank update}(\vtheta^{t}) $ \\
\textbf{Output:} compressed weights $\hat{\vtheta} = \vtheta^T$ 
\end{algorithmic}
\end{algorithm}

\subsection{Dampening}
\label{sec:dampening}

In practice, we dampen the $\mG$ and $\mA$ matrices by adding a diagonal term $\mG + \lambda_G \mI$ and $\mA + \lambda_A \mI$. In our experiments, we found that values in the range [0.01, 0.1] multiplied by mean diagonal terms generally works well. We follow \citep{frantar2023sparsegpt} and always use $\lambda_A{=}0.01\text{diag}(\mA)$ to be consistent with prior work and allow for a fair comparison with baselines. Further, we use $\lambda_G{=}0.1\text{diag}(\mG)$ for structured experiments and $\lambda_G{=}0.01\text{diag}(\mG)$ in semi-structured and unstructured experiments. %If the same weighting is used for $\lambda_G$ and $\lambda_A$, the dampening can be interpreted as a prior variance from a Bayesian perspective. %Similar to other observations made in optimisation \citep{martens2015optimizing}, we observe that different values for $\lambda_G$ and $\lambda_A$ can sometimes improve performance.

% \subsection{Structured thresholding}
% \label{sec:dynamic-thresholding-details}

% \subsection{Controlling the maximum number of correlated weights}
% \label{sec:maximum-amount-of-correlated-weights}

\newpage
\section{Downstream task performance}
\label{sec:downstream-tasks}

We also evaluate our method on downstream tasks as perplexity metrics do not necessarily correlate with downstream performance. Further, we also repeat this experiment using the C4 dataset as reference data for compression, as this is used in prior work \citep{frantar2023sparsegpt} and as this can be regarded a more general reference dataset. In \cref{tab:downstream-wikitext2,tab:downstream-c4} we report 0-shot test performance of structured pruning for LLM surgeon and K-OBD baseline.

\vspace{-1em}

\begin{table}[h]
\centering
\caption{Downstream task performance using Wikitext-2 for pruning.}
\resizebox{\linewidth}{!}{
    \begin{tabular}{l|l|llllllllllll}
        \textbf{Structured pruning} \\
        (with wikitext-2) & \textbf{Model size} & \textbf{wikitext word ppl} & \textbf{boolq} & \textbf{piqa} & \textbf{hallaswag} & \textbf{winogrande} & \textbf{arc\_easy} & \textbf{arc\_challenge} & \textbf{openbookq} & \textbf{copa} & \textbf{lambada\_openai} & \textbf{wsc273} & \textbf{AVERAGE wikitext2} \\ \hline
        \textbf{Dense baseline} & 100\% & 9.24 & 77.74 & 79.11 & 75.99 & 69.14 & 74.58 & 46.25 & 44.20 & 86.00 & 73.92 & 85.71 & 71.26 \\ \hline
        \textbf{LLM Surgeon} (\textbf{ours}) & 90\% & 9.63 & 76.21 & 78.56 & 75.39 & 67.64 & 74.12 & 46.50 & 43.60 & 85.00 & 72.64 & 84.98 & 70.46 \\ \hline
        & 80\% & 12.16 & 72.97 & 77.09 & 71.30 & 66.30 & 71.36 & 41.89 & 41.80 & 87.00 & 56.43 & 80.22 & 66.66 \\ \hline
        & 70\% & 16.91 & 61.25 & 73.56 & 60.72 & 61.09 & 63.09 & 36.69 & 38.80 & 81.00 & 28.33 & 76.56 & 58.11 \\ \hline
        & 60\% & 25.15 & 44.98 & 69.26 & 48.04 & 54.38 & 52.31 & 30.29 & 36.80 & 78.00 & 11.72 & 68.50 & 49.43 \\ \hline
        & 50\% & 43.68 & 39.60 & 64.36 & 40.29 & 52.57 & 44.91 & 26.28 & 30.80 & 74.00 & 6.52 & 61.54 & 44.09 \\ \hline
        \textbf{K-OBD} & 90\% & 9.89 & 76.67 & 78.02 & 74.80 & 68.11 & 75.17 & 46.33 & 44.60 & 86.00 & 72.71 & 82.78 & 70.52 \\ \hline
        & 80\% & 17.62 & 74.34 & 75.24 & 67.85 & 64.64 & 63.80 & 40.27 & 41.60 & 83.00 & 30.23 & 82.42 & 62.34 \\ \hline
        & 70\% & 32.72 & 65.29 & 71.82 & 53.07 & 56.83 & 51.05 & 33.11 & 37.80 & 79.00 & 12.21 & 70.70 & 53.09 \\ \hline
        & 60\% & 68.63 & 60.80 & 65.67 & 43.99 & 53.20 & 41.79 & 28.50 & 34.00 & 75.00 & 7.04 & 60.44 & 47.04 \\ \hline
        & 50\% & 136.33 & 61.56 & 60.66 & 36.84 & 53.04 & 36.11 & 26.71 & 33.00 & 72.00 & 4.70 & 61.17 & 44.58 \\ \hline
    \end{tabular}
    }
    \label{tab:downstream-wikitext2}
\end{table}

\vspace{-2em}

\begin{table}[h]
\centering
\caption{Downstream task performance using C4 for pruning.}
\resizebox{\linewidth}{!}{
    \begin{tabular}{l|l|llllllllllll}
    \hline
        \textbf{Structured pruning} \\
        (with C4) & \textbf{Model size} & \textbf{wikitext word ppl} & \textbf{boolq} & \textbf{piqa} & \textbf{hallaswag} & \textbf{winogrande} & \textbf{arc\_easy} & \textbf{arc\_challenge} & \textbf{openbookq} & \textbf{copa} & \textbf{lambada\_openai} & \textbf{wsc273} & \textbf{AVERAGE wikitext2} \\ \hline
        \textbf{Dense baseline} & 100\% & 9.24 & 77.74 & 79.11 & 75.99 & 69.14 & 74.58 & 46.25 & 44.20 & 86.00 & 73.92 & 85.71 & 71.26 \\ \hline
        \textbf{LLM Surgeon} (\textbf{ours}) & 90\% & 9.90 & 77.03 & 78.45 & 74.95 & 68.27 & 73.19 & 45.99 & 44.60 & 84.00 & 72.81 & 82.78 & 70.21 \\ \hline
        & 80\% & 14.42 & 75.60 & 76.82 & 69.71 & 63.85 & 70.29 & 41.30 & 42.80 & 87.00 & 45.53 & 82.42 & 65.53 \\ \hline
        & 70\% & 25.16 & 66.39 & 72.85 & 58.11 & 56.83 & 62.16 & 34.47 & 38.40 & 80.00 & 22.69 & 69.96 & 56.19 \\ \hline
        & 60\% & 45.35 & 62.48 & 68.93 & 48.10 & 55.64 & 51.56 & 27.99 & 35.20 & 70.00 & 12.56 & 61.54 & 49.40 \\ \hline
        & 50\% & 77.30 & 62.60 & 65.02 & 41.70 & 54.22 & 42.55 & 24.23 & 31.20 & 71.00 & 7.26 & 60.44 & 46.02 \\ \hline
        \textbf{K-OBD} & 90\% & 10.59 & 75.47 & 78.18 & 73.61 & 66.46 & 72.52 & 44.37 & 43.60 & 87.00 & 71.22 & 82.42 & 69.48 \\ \hline
        & 80\% & 20.12 & 73.36 & 75.14 & 66.11 & 62.43 & 62.84 & 38.23 & 41.00 & 86.00 & 21.50 & 78.39 & 60.50 \\ \hline
        & 70\% & 56.92 & 63.30 & 68.44 & 52.31 & 55.64 & 46.72 & 31.31 & 34.60 & 77.00 & 5.69 & 68.86 & 50.39 \\ \hline
        & 60\% & 112.85 & 62.23 & 64.47 & 46.36 & 52.17 & 40.53 & 29.52 & 32.40 & 72.00 & 2.91 & 63.00 & 46.56 \\ \hline
        & 50\% & 272.16 & 62.42 & 61.70 & 38.47 & 50.43 & 33.29 & 26.96 & 31.80 & 65.00 & 0.91 & 59.34 & 43.03 \\ \hline 
    \end{tabular}
}
\label{tab:downstream-c4}
\end{table}

We find that our method not only performs well in terms of test perplexity but also correlates well with downstream performance, outperforming the baselines on these downstream tasks.

\section{Additional experiments on Llama-v2 13B.}

To assess performance on larger 13B parameter models, we also report structured compression on the Llama-v2 13B model and evaluate downstream task performance. Test perplexities (lower is better) can be found in \cref{tab:13b-ppl} below:

\begin{table}[h]
\centering
    \caption{Pruning Llama-v2 13B model.}
    \begin{tabular}{l|c|ccccc}
& \textbf{Baseline} & \multicolumn{5}{c}{\textbf{Pruned model sizes}} \\
& Dense 100\% & 90\% & 80\% & 70\% & 60\% & 50\% \\
\hline
\textbf{K-OBD} & 4.547 & 4.908 & 6.294 & 10.08 & 13.06 & 16.06 \\
\textbf{LLM Surgeon} & 4.547 & 4.692 & 5.286 & 6.207 & 7.245 & 9.428 \\
    \end{tabular}
    \label{tab:13b-ppl}
\end{table}

as well as evaluated results on downstream benchmarks (higher is better) in \cref{tab:downstream-13B} below.

\begin{table}[h!]
\caption{Downstream task performance after pruning large Llama-v2 13B model.}
\resizebox{\linewidth}{!}{
    \begin{tabular}{l|l|llllllllllll}
    \hline
        \textbf{Llama-v2 13B} & \textbf{Model size} & \textbf{wikitext word ppl} & \textbf{boolq} & \textbf{piqa} & \textbf{hallaswag} & \textbf{winogrande} & \textbf{arc\_easy} & \textbf{arc\_challenge} & \textbf{openbookq} & \textbf{copa} & \textbf{lambada\_openai} & \textbf{wsc273} & \textbf{AVERAGE wikitext2} \\ \hline
        \textbf{Dense baseline} & 100\% & 8.23 & 80.52\% & 80.52\% & 79.38\% & 72.14\% & 77.53\% & 49.23\% & 45.20\% & 90.00\% & 76.77\% & 89.38\% & 74.07\% \\ \hline
        \textbf{LLM Surgeon} (\textbf{ours}) & 90\% & 8.57 & 81.07\% & 79.87\% & 79.24\% & 72.38\% & 76.30\% & 49.91\% & 47.20\% & 92.00\% & 75.47\% & 89.38\% & 74.28\% \\ \hline
        & 80\% & 10.08 & 80.86\% & 79.00\% & 77.09\% & 70.56\% & 75.93\% & 46.76\% & 46.80\% & 90.00\% & 67.79\% & 86.45\% & 72.12\% \\ \hline
        & 70\% & 12.74 & 74.50\% & 76.50\% & 71.52\% & 68.67\% & 69.74\% & 40.27\% & 45.00\% & 91.00\% & 54.40\% & 83.52\% & 67.51\% \\ \hline
        & 60\% & 16.00 & 64.62\% & 73.01\% & 65.04\% & 65.75\% & 63.80\% & 37.12\% & 39.60\% & 90.00\% & 44.50\% & 81.32\% & 62.48\% \\ \hline
        & 50\% & 23.75 & 65.66\% & 68.77\% & 56.19\% & 63.22\% & 56.19\% & 31.83\% & 36.60\% & 85.00\% & 35.16\% & 77.29\% & 57.59\% \\ \hline
        \textbf{K-OBD} & 90\% & 8.79 & 81.31\% & 79.76\% & 79.12\% & 72.22\% & 76.94\% & 47.95\% & 47.80\% & 91.00\% & 75.26\% & 88.64\% & 74.00\% \\ \hline
        & 80\% & 11.79 & 80.80\% & 79.16\% & 76.80\% & 70.56\% & 73.74\% & 46.93\% & 48.60\% & 88.00\% & 58.99\% & 87.55\% & 71.11\% \\ \hline
        & 70\% & 20.00 & 66.76\% & 74.43\% & 64.18\% & 64.96\% & 56.23\% & 36.01\% & 39.00\% & 88.00\% & 38.54\% & 79.49\% & 60.76\% \\ \hline
        & 60\% & 27.74 & 55.66\% & 70.24\% & 55.52\% & 60.46\% & 49.62\% & 32.68\% & 35.80\% & 80.00\% & 30.06\% & 73.63\% & 54.37\% \\ \hline
        & 50\% & 37.38 & 59.79\% & 66.54\% & 48.39\% & 57.46\% & 46.59\% & 30.72\% & 34.00\% & 77.00\% & 24.61\% & 69.96\% & 51.50\% \\ \hline
    \end{tabular}
}
\label{tab:downstream-13B}
\end{table}

We find that LLM Surgeon also outperforms baselines on existing Llama-v2 13B models. We stress that these results are obtained on structured pruning of rows and columns, which are regarded the hardest and most constrained pruning structure. Yet, we can compress Llama 13B by 20\% with less than 2\% drop in downstream task performance. It also significantly outperforms the baseline for all compression rates, both in terms of test perplexity and downstream task performance.

\newpage
\section{Ablations}

\subsection{Shots}
\label{sec:ablation-shots}

\begin{table}[h]
\caption{Ablation of shot counts $T$ for structured LLM Surgeon compressing OPT-1.3b model.}
\centering
\begin{tabular}{c | c c | c c | c c }
\bf{Target size} & Shots $T$ & wikitext-2 PPL & Shots $T$ & wikitext-2 PPL & Shots $T$ & wikitext-2 PPL \\
\hline
90\% & 6 & 14.70 & 8 & 14.70 & 10 & 14.72 \\
80\% & 12 & 15.14 & 16 & 15.12 & 20 & 15.08 \\
70\% & 18 & 16.21 & 24 & 16.24 & 30 & 16.23 \\
60\% & 24 & 18.53 & 32 & 18.45 & 40 & 18.49 \\
50\% & 30 & 23.32 & 40 & 22.95 & 50 & \bf{22.68}
\end{tabular}
\end{table}

% %\newpage
\subsection{Task-specific compression}
\label{sec:results-tasks}

\begin{wraptable}{h}{70mm}
\vspace{-2.3em}
\caption{Cross-task performance and mask equivalences of 50\% compressed OPT-125m model using structured LLM Surgeon on language subsets.}
\label{tab:results-task-specific}
\begin{center}
\resizebox{\linewidth}{!}{
\begin{tabular}{c|cccc|cccc}
& \multicolumn{4}{c}{\bf evaluation dataset }
& \multicolumn{4}{c}{\bf mask equivalence (\%)}
\\
\bf{target}
& EN & FR & DE & IT
& EN & FR & DE & IT
\\
\hline
%LLM Surgeon (\textbf{ours}) on OPT-125m & -
%-
% & \bf{27.65} & \bf{22.63} & \bf{22.21} & \bf{26.7}
%& 1.0 & 1.0 & 1.0 & 1.0
%& & & &
%\\
%\hline
Pretrained & \bf{27.66} & \bf{22.54} & \bf{24.32} & \bf{27.66} \\
\hline
% Target size: 50\%
EN
% & \bf{50.52} & 160.3 & 179.3 & 173.5
& \bf{47.46} & 172.9 & 181.1 & 169.1 & 1.00 & 0.74 & 0.70 & 0.72 
\\
FR
%& 118.8 & \bf{28.16} & 32.33 & 32.02 
& 113.4 & \bf{28.44} & 35.02 & 34.90 & 0.74 & 1.00 & 0.87 & 0.90% <-- with right damp param
\\
DE
%& 143.7 & 35.59 & \bf{26.59} & 41.58
& 142.1 & 35.15 & \bf{27.49} & 38.49 & 0.70 & 0.87 & 1.00 & 0.87 
\\
IT
%& 129.8 & 31.88 & 32.87 & \bf{32.02}
& 123.7 & 31.85 & 33.78 & \bf{30.58} & 0.72 & 0.90 & 0.87 & 1.00

\\
\hline
\end{tabular}
}
\end{center}
\vspace{-1em}
\end{wraptable}

LLM Surgeon uses data to find a compressed model that has the least negative impact on final test performance. In this section, we explore the extent to which the method can use data to compress specifically to the task at hand. We do so by comparing test performance and equivalences between resulting pruning masks for different language modeling languages: English (EN/wikitext-2), French (FR) and Italian (IT) and the German (DE). We consider 50\% unstructured compression using LLM Surgeon with correlated weight updates. For each compressed model, we compare performance on all languages and compare the equivalences between resulting pruning masks (details in \cref{sec:mask-equivalence}), and report results in \cref{tab:results-task-specific}. Like other methods that use data for compression \citep{hassibi1992second,frantar2023sparsegpt,wang2019eigendamage}, we expect to see some correlation between the data used for training and data with good test performance, which is reflected in both test performance and masks. It is important to note that the final performance after compression will depend on the quality of the used dataset for compression. Further, the experiment demonstrates that the method can be used for task-specific compression tailored towards the data used for compression and generalises to high test performance on the associated test data.

\newpage
\section{On fair comparison}
\label{sec:fair-comparison}

All results in this work (including the SparseGPT) were trained on Wikitext-2 for fair comparison. To do so, we used the same dataloader and evaluation script as the official SparseGPT repo and reran all SparseGPT results to be trained on Wikitext-2. In some cases, this resulted in better scores for the SparseGPT baseline compared to the C4-trained results reported in the original SparseGPT paper. Yet, we find that our method using improved curvature estimates still outperformed the baselines in terms of final test performance.

\section{Computational performance}
\label{sec:time-performance}

We report computational cost in terms of pruning time in \cref{tab:timing-performance} and GPU memory in \cref{tab:memory-performance}.

\begin{table}[h]
\caption{Time performance.}
\begin{center}
\resizebox{\linewidth}{!}{
\begin{tabular}{ll|l|lllll}
& & & \multicolumn{5}{c}{Test performance} \\
Runtime	& Network & Time & PPL 90\% & PPL 80\% & PPL 70\% & PPL 60\% & PPL 50\% \\
\hline
Unstructured baseline (SparseGPT) & Llama-v2 7B & $<$5m & 5.49 & 5.58 & 5.71 & 5.94 & 6.51 \\
Unstructured LLM Surgeon (\textbf{ours}) & Llama-v2 7B & 2d8h16m & 5.13& 5.20 & 5.36 & 5.66 & 6.08 \\
\hline
Structured baseline (K-OBD) & Llama-v2 7B & 16h58m & 5.48 & 9.14 & 15.43 & 28.03 & 46.64 \\
Structured LLM Surgeon (\textbf{ours}) & Llama-v2 7B & 17h08m & 5.25 & 6.18 & 7.83 & 10.39 & 15.38 \\
Structured baseline (K-OBD) & Llama-v2 13B & 1d6h5m & 4.908 & 6.294 & 10.08 & 13.06 & 16.06 \\
Structured LLM Surgeon (\textbf{ours}) & Llama-v2 13B & 1d9h26m & 4.692 & 5.286 & 6.207 & 7.245 & 9.428 
\end{tabular}
}
\end{center}
\label{tab:timing-performance}
\vspace{-1.0em}
\end{table}

Our method is most efficient for structured pruning, but it must be noted that engineering efforts may further improve speed for unstructured pruning. The focus of the paper is structured pruning, on which we achieve state-of-the-art compression rates. Importantly, compression of LLMs only needs to happen once after which a pruned model can be deployed infinitely many times without further cost. This motivates our method which takes longer to run but reaches better final test performance. 

\begin{table}[h]
\caption{Memory performance.}
\begin{center}
% \resizebox{\linewidth}{!}{
\begin{tabular}{ll| ll ll }
Network & SparseGPT (baseline) & Unstructured LLM-Surgeon (ours) \\
Llama-7B & $<$5m / 1 GPU (32GB) & 2d8h16m / 4xH100 80 GB \\
\hline
& K-OBD (baseline) & Structured LLM-Surgeon (ours) \\
Llama-7B & 16h58m / 4xH100 80 GB & 17h08m / 4xH100 80 GB \\
Llama-13B & 1d6h5m / 8xH100 80 GB & 1d9h26m / 8xH100 80 GB \\
\end{tabular}
% }
\end{center}
\label{tab:memory-performance}
\vspace{-1.0em}
\end{table}

We argue that differences in the performance and the runtime of pruning methods can largely be attributed to underlying assumptions on correlations between weights. Notably, algorithms that consider few correlations, sometimes to the extent of completely disregarding all gradient information, can result in very fast pruning algorithms for unstructured and semi-structured pruning but are often not flexible enough to perform structured pruning of rows and columns. Examples of such lightweight algorithms for LLMs are \citep{sun2023simple} and SparseGPT \citep{frantar2023sparsegpt}, as can also be observed from \cref{tab:timing-performance}. Our approach makes less strong assumptions on the curvature of the loss and as a result outperforms all baselines on all unstructured, semi-structured and structured pruning. Further, the improved curvature is also eligible for dynamic allocation of weight removal and improved correlated weight updates. In practice, we always recommend using our method for structured pruning. For unstructured and semi-structured pruning, we note an important trade-off between the desired final test accuracy and the available computational budget. Here, our proposed method can achieve the highest final model performance but requires more computational resources and takes longer to run. It should be noted that pruning only needs to happen once after which a model can be deployed infinitely many times this time, which dependent on the available computational resources can also legitimise spending additional pruning time even if this is much higher compared to other algorithms in relative terms. In absolute terms, the use of multiple large GPUs is common practice in the field of large language models and many more GPUs are typically used to train and deploy large language models. Moreover, the curvature approximation is naively amenable to data parallelism in case further speed-ups or larger models are required. We hope this provides context and emphasises the trade-off between performance and compute in practice.

\newpage
\section{Extending curvature estimates}
\label{sec:extending-curvature-estimates}

Instead of using a single Kronecker product, we might consider improving the approximation through a sum of multiple Kronecker factors:
\begin{align}
\mF \approx \widetilde{\mF} = \mG_1 \otimes \mA_1 + \mG_2 \otimes \mA_2
\end{align}
This last appendix deals with the question how one may computationally find such approximations and how to utilise them in the neural network pruning framework.

\subsection{Nearest Kronecker product or sum of Kronecker products}

Instead of assuming independence of activations and derivatives as in \cref{sec:estimating-curvature}, following the classic KFAC of \citep{martens2015optimizing}, we might want to find the \textit{nearest Kronecker product} approximation $\mF \approx \widetilde{\mG} \otimes \widetilde{\mA}$ that is closest to the Fisher in terms of the Frobenius norm:
\begin{align}
\label{eq:nkp-problem}
\widetilde{\mG}_l, \widetilde{\mA}_l
&= \argmin_{\mG_l, \mA_l} || \mF_l - \mG_l \otimes \mA_l ||_F 
\end{align}
Finding the nearest sum of Kronecker factors can be rephrased as a classic eigenvalue problem of finding the nearest rank-1 matrix. \cite{golub2013matrix}. 
\begin{align}
|| \mF - \tilde{\mG} \otimes \tilde{\mA} ||_F
\hspace{1em} \equiv \hspace{1em}
|| \mathcal{R}(\mF) - \text{vec}(\widetilde{\mG}) \text{vec}(\widetilde{\mA})^T ||_F
\end{align}

\paragraph{Power method and deflation}
After considering the reshaping, we can use power iterations to solve for and find the nearest Kronecker factors $\mG_1, \mA_1 = \text{solve}(\mF)$.
\begin{align}
\begin{split}
&\text{Find with power iterations:}\\
\widetilde{G}_1, \widetilde{A}_1 &= \text{solve}(\mF) = \argmin_{\mG, \mA} || \mF - \mG \otimes \mA ||_F
\end{split}
\hspace{1em}
\begin{split}
&\text{Deflation:}\\
\widetilde{G}_r, \widetilde{A}_r &= \text{solve}(\mF - \sum\nolimits_{r'=1}^{r-1}(\widetilde{\mG}_{r'} \otimes \widetilde{\mA}_{r'}) ) \nonumber
\end{split}
\end{align}
A more extensive description of the power method $\text{solve}(\cdot)$ can be found in \cref{alg:pseudocode-power-method}. At the start of the algorithm, we initialise power iterations as vector with one's $\mathbf{1} = \begin{bmatrix} 1 & 1 & \ldots & 1 \end{bmatrix}$. After each shot we can initialise the vector as the final estimate found during the previous shot. %We use 10 power iterations at the first shot, but use only a single power iteration to update the curvature estimate in subsequent shots.

\begin{algorithm}[h]
\caption{Kronecker power method. Finds $\widetilde{\mG}, \widetilde{\mA}$ nearest Kronecker product $|| \mF - \widetilde{\mG} \otimes \widetilde{\mA} ||_F$.} \label{alg:pseudocode-power-method}
\begin{algorithmic}
\Require Initialise $\widetilde{\vg}^0 {=} \mathbf{1}, \widetilde{\va}^0 {=} \mathbf{1}$ (or using estimates of previous shot). 
\Require Set iterations $I$ (or $I{=}1$ if using estimates from previous shot)
\Ensure $\widetilde{\mG}, \widetilde{\mA}$
\For{iteration $i$ \text{in} [1, 2, \ldots, $I$]} \\
    \hspace{1em} \textbf{Compute:} $\widetilde{\vg}^i = \frac{ \mathcal{R}(\widetilde{\mF}) \widetilde{\va}^{i-1} }{ || \mathcal{R}(\widetilde{\mF}) \widetilde{\va}^{i-1} ||_2 } $ \text{, with} $\mathcal{R}(\widetilde{\mF}) \widetilde{\va}^{i-1} = \frac{1}{N} \sum_{n=1}^N \va_{n}^T \widetilde{\mA}^{i-1} \va_{n} \text{vec}( \vg_n \vg_n^T) $ \\
    \hspace{1em} \textbf{Compute:} $\widetilde{\va}^i = \frac{ \mathcal{R}(\widetilde{\mF})^T \widetilde{\vg}^{i} }{ || \mathcal{R}(\widetilde{\mF})^T \widetilde{\vg}^{i} ||_2 } $ \text{, with} $\mathcal{R}(\widetilde{\mF})^T \widetilde{\vg}^{i} = \frac{1}{N} \sum_{n=1}^N \vg_n^T \widetilde{\mG}^i \vg_n \text{vec}(\va_n \va_n^T) $\\
    \hspace{1em} \textbf{Compute:} $\sigma^i = || \widetilde{\va}^i ||_2 $
\EndFor \\
\textbf{Return:} $\widetilde{\mG} = \sqrt{\sigma^i} \text{mat}(\widetilde{\vg}), \widetilde{\mA} = \sqrt{\sigma^i} \text{mat}(\widetilde{\va})$.
\end{algorithmic}
\end{algorithm}

\begin{figure}[b]
    \centering
    \resizebox{\linewidth}{!}{
    \includegraphics[]{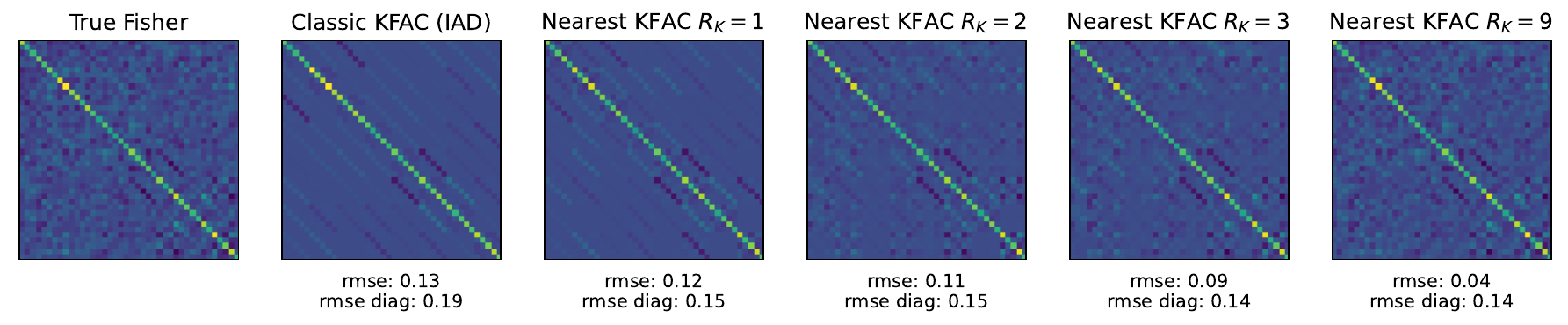}
    }
    \vspace{-2em}
    \caption{Example illustration of nearest Kronecker factor approximations $\widetilde{\mF} {\approx} \sum_{r=1}^{R_K} \mG_i \otimes \mA_i$, compared to classical KFAC with the IAD assumption. Larger $R_K$ yields better approximations to the true Fisher $\mF$ for larger $R_K$, as measured by the root mean squared error (rmse).}
    \vspace{-1.0em}
\end{figure}

%\newpage
\subsection{Extended curvature approximations}

For classic KFAC with IAD or $R_K{=}1$ nearest Kronecker approximations of the form $\widetilde{\mF} = \mG \otimes \mA$, the inverse simply becomes $(\mG \otimes \mA)^{-1} = \mG^{-1} \otimes \mA^{-1}$. Unfortunately, we can not use this famous inverse identity for sum of Kronecker factors, which is why we fall back on eigendecompositions $\mG = \mE_1 \mS_1 \mE_1^T$ and $\mA = \mE_2 \mS_2 \mE_2^T$, allowing us to decompose the Fisher into:
\begin{align}
\widetilde{\mF} = \mK_1 \mS_1 \mK_1^T \otimes \mK_2 \mS_2 \mK_2^T = (\mK_1 \otimes \mK_2) (\mI \otimes \mI + \mS_1 \otimes \mS_2) (\mK_1^T \otimes \mK_2^T)
\end{align}
where specific $\mK_1$ and $\mK_2$ can be found in App. B of \cite{martens2015optimizing}, which we closely followed in our derivations. Because $\mK_1$ and $\mK_2$ are orthogonal and $\mS_1$ and $\mS_2$ diagonal, the inverse Fisher becomes:
\begin{align}
\widetilde{\mF}^{-1} = (\mK_1 \otimes \mK_2) (\mI \otimes \mI + \mS_1 \otimes \mS_2)^{-1} (\mK_1^T \otimes \mK_2^T)
\end{align}

In the context of neural network training, the problem gets slightly harder since we want to incrementally construct estimates $\widetilde{\mG}_i$ and $\widetilde{\mA}_i$ from individual samples $\va_{l, n}, \vg_{l, n}$ that make up $\mF$, without having to simultaneously store more than a single or batch of input activations $\va_{l, n}$ or output gradients $\vg_{l, n}$ in memory. Although this \textit{online Kronecker-product principal component analysis} problem largely remains an open research problem, we our approach closely follows the recent work by \citep{koroko2022efficient} that uses similar approximations in the context of optimisation. A sum of multiple $R_K{>}1$ Kronecker factors will yield closer approximations, but also linearly increase memory requirements with higher $R_K$ and makes inverting $\mF^{-1}$ considerably more difficult. %Yet, we derive unstructured, semi-structured and structured pruning for these extended family of Fisher approximations in \cref{sec:svd-pruning} through use of eigendecompositions.

\paragraph{Formulas to compute cost and weight updates.} For sum of Kronecker factors, we find that the constrained optimization solution of for costs $\Delta \mathcal{L}$ \cref{eq:general-cost} and weight updates $\Delta \vtheta$ \cref{eq:general-update} become the following inner-product and matrix-vector product:
\begin{align}
\mathcal{L}_k = \frac{1}{2} \langle \widebar{\vtheta^*}, \mU \widebar{\vtheta^*} \rangle &= (\widebar{\vtheta^*})^T \mU (\widebar{\vtheta^*}) \in \R \label{eq:general-framework-loss} \\
\Delta \vtheta = \widetilde{\mF}^{-1} \mE_K^T \vu &= \mK_1 \left( \widebar{\mK}_1^T \mU \widebar{\mK}_2 \oslash \left[ \mathbf{1}\mathbf{1}^T + \vs_1 \vs_2^T \right] \right) \mK_2^T \in \R^{RC} \label{eq:general-framework-update}
\end{align}
with at the heart of it all a matrix $\mU = [\mE_K \mF^{-1} \mE_K^T]^{-1}$ that captures correlations between weights:
\begin{align}
\label{eq:formula-u}
\mU &= \left[ \mE_K \Big(\mK_1 \otimes \mK_2\Big) \Big( \mI \otimes \mI + \mS_1 \otimes \mS_2 \Big)^{-1} \Big( \mK_1^T \otimes \mK_2^T \Big) \mE_K^T \right]^{-1}
\end{align}
where $(\mI \otimes \mI + \mS_1 \otimes \mS_2)$ is diagonal and the inverse can thus be computed element-wise. The remaining inverse is of size $K \times K$, for $K$ correlated weights.

\paragraph{Note on sum of Kronecker factors} Experimentally, we did not find a benefit in performance when using a sum of two nearest Kronecker factor approximation, or found it too slow. Therefore, we focus in the main text on LLM Surgeon with fast single Kronecker product KFAC approximation to approximate the loss landsscape curvature. Nevertheless, we choose to include this appendix as we believe could prove useful in other contexts or inspire future work that aim to further improve the quality of curvature approximations.

\newpage
\section{Code}

Code is available at: \href{https://github.com/Qualcomm-AI-research/llm-surgeon}{https://github.com/Qualcomm-AI-research/llm-surgeon}.

\end{document}